\documentclass{article} 
\usepackage{iclr2021_conference,times}

\usepackage{listings}
\usepackage[utf8]{inputenc}
\usepackage[english]{babel}
\usepackage{natbib}
\setcitestyle{authoryear,open={(},close={)}}
\usepackage{wrapfig}

\usepackage{hyperref}
\usepackage{url}

\usepackage{times}  
\usepackage{graphicx} 

\usepackage{fixltx2e}
\usepackage{subcaption}
\usepackage[cmex10]{amsmath}
\usepackage{amssymb}
\usepackage{amsfonts}
\usepackage{amsmath}
\usepackage{supertabular}
\usepackage{times}
\usepackage{amsthm}
\usepackage{floatrow}
\usepackage{tikz}
\usepackage{float}
\usepackage{mathtools}
\usepackage{textcomp}
\usepackage{multirow}
\usepackage{algpseudocode}
\usepackage[ruled,vlined]{algorithm2e}
\usepackage{adjustbox}
\usepackage[justification=centering, font=small]{caption} 
\usepackage{epstopdf}
\usepackage{tabularx}
\usepackage{nccmath}
\usepackage{mathtools}
\usepackage{easyReview}

\theoremstyle{plain}
\newtheorem{thm}{Theorem}
\newtheorem{defn}{Definition}
 
\newtheorem{lemma}{Lemma}

\newtheorem{corr}{Corollary}

\DeclareMathOperator*{\argmin}{arg\,min}

\newcommand{\vect}[1]{\boldsymbol{#1}}

\newcommand{\cT}{\vect{\mathcal{T}}} 
\newcommand{\cX}{\vect{\mathcal{X}} } %
\newcommand{\cY}{\vect{\mathcal{Y}} } %
\newcommand{\cD}{\vect{\mathcal{D}} }
\newcommand{\mR}{\vect{\mathbb{R}} }

\graphicspath{{./Figures/}}

\title{Meta Continual Learning via Dynamic Programming}

\author{R.~Krishnan$^{1}$ and Prasanna~Balaprakash$^{1,2}$  \\
    $\text{ }^{1}$Mathematics and Computer Science Division\\
    $\text{ }^{2}$Leadership Computing Facility\\ 
    Argonne National Laboratory\\
  \textit{kraghavan,pbalapra@anl.gov}} 
\iclrfinalcopy 
\begin{document}
\maketitle
\begin{abstract}
Meta continual learning algorithms seek to train a model when faced with similar tasks observed in a sequential manner. Despite promising methodological advancements, there is a lack of theoretical frameworks that enable analysis of learning challenges such as generalization and catastrophic forgetting. To that end, we develop a new theoretical approach for meta continual learning~(MCL) where we mathematically model the learning dynamics using dynamic programming, and we establish conditions of optimality for the MCL problem.  Moreover, using the theoretical framework, we derive a new dynamic-programming-based MCL method that adopts stochastic-gradient-driven alternating optimization to balance generalization and catastrophic forgetting. We show that, on MCL benchmark data sets, our theoretically grounded method achieves accuracy better than or comparable to that of existing state-of-the-art methods.
\end{abstract}
\section{Introduction}
\begin{wrapfigure}[18]{r}{0.5\textwidth}
	\includegraphics[width =\linewidth]{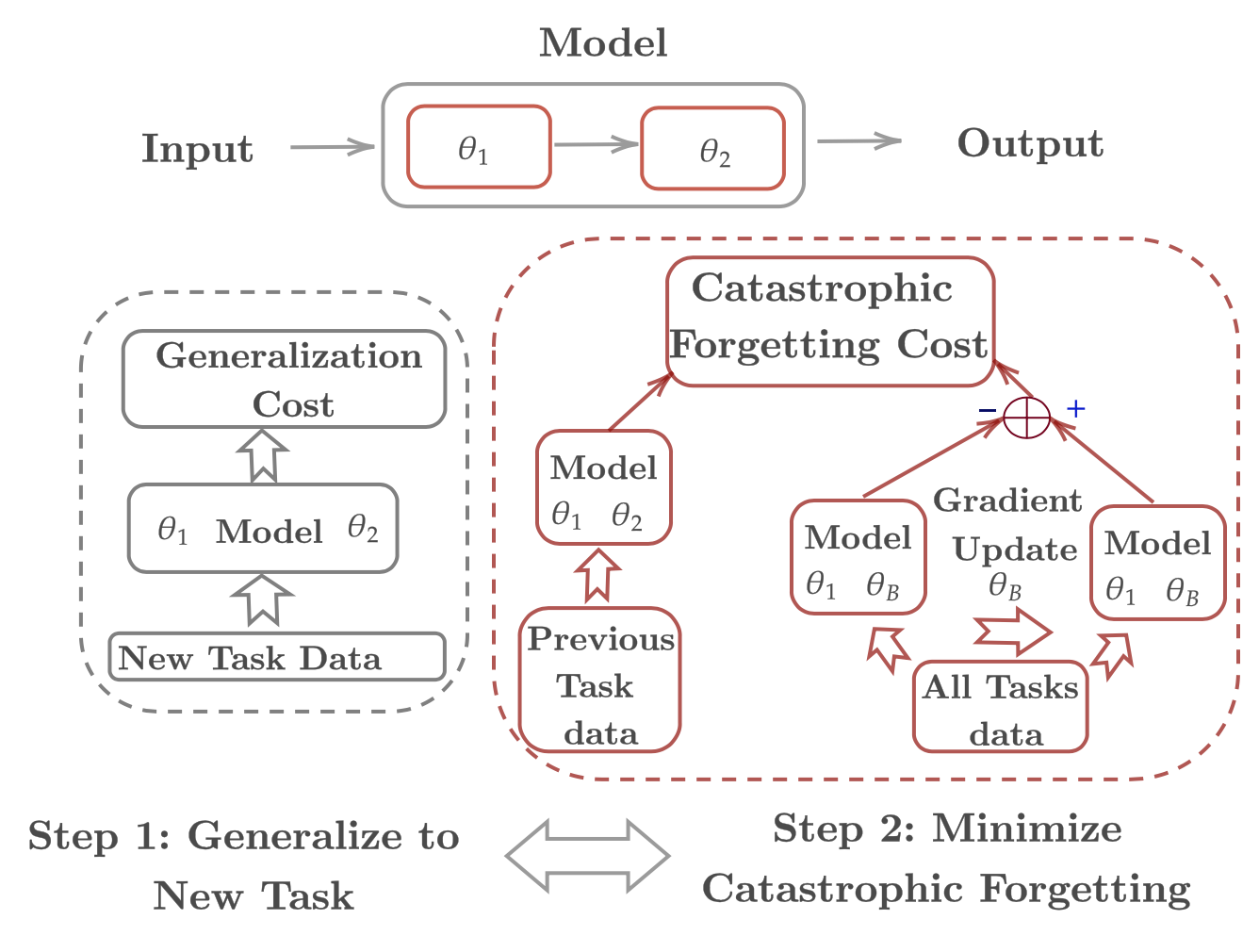} 
	\caption{Illustration of DPMCL to learn parameters $\hat{\vect{\theta}}_{1}, \hat{\vect{\theta}}_{2}$ (the copy of $\hat{\vect{\theta}}_{2}$ is $\hat{\vect{\theta}}_{B}).$
	}
	\label{fig:methodology}
\end{wrapfigure}
The central theme of meta continual learning (MCL) is to learn on similar tasks revealed sequentially. In this process, two fundamental challenges must be addressed: catastrophic forgetting of the previous tasks and generalization to new tasks \citep{finn2017model,javed2019meta}. In order to address these challenges, several approaches \citep{beaulieu2020learning,finn2017model,javed2019meta} have been proposed in the literature that build on the second-order, derivative-driven approach introduced in \citet{finn2017model}.

Despite the promising prior methodological advancements, existing MCL methods suffer from three key issues: (1) there is a lack of a theoretical framework to systematically design and analyze MCL methods; (2) data samples representing the complete task distribution must be known in advance~\citep{finn2017model, javed2019meta, beaulieu2020learning}, often, an impractical requirement in real-world environments as the tasks are observed sequentially; and (3) the use of fixed representations \citep{javed2019meta, beaulieu2020learning} limits the ability to handle significant changes in the input data distribution, as demonstrated in \citet{Caccia2020OnlineFA}. We focus on a supervised learning paradigm within MCL, and our key contributions are (1) a dynamic-programming-based theoretical framework for MCL and (2) a theoretically grounded MCL approach with convergence properties that compare favorably with the existing MCL methods.

{\bf Dynamic-programming-based theoretical framework for MCL:} 
In our approach, the problem is first posed as the minimization of a cost function that is integrated over the lifetime of the model. Nevertheless, at any time $t$, the future tasks are not available, and the integral calculation becomes intractable. Therefore, we use the Bellman's principle of optimality \citep{bellman2015adaptive} to recast the MCL problem to minimize the sum of catastrophic forgetting cost on the previous tasks and generalization cost on the new task. Next, we theoretically analyze the impact of these costs on the MCL problem using tools from the optimal control literature~\citep{lewis2012optimal}. Furthermore, we demonstrate that the MCL approaches proposed in \citep{finn2017model, beaulieu2020learning, javed2019meta} can be derived from the proposed framework.

{\bf Theoretically grounded MCL approach:} We derive a theoretically grounded dynamic programming-based meta continual learning (DPMCL) approach. In our approach, the generalization cost is computed by training and evaluating the model on given new task data. The catastrophic forgetting cost is computed by evaluating  the model on the task memory (previous tasks) after the model is trained on the new task. We alternately minimize the generalization and catastrophic forgetting costs for a predefined number of iterations to achieve a balance between the two. Our approach is illustrated in Fig. \ref{fig:methodology}. We analyze the performance of the DPMCL approach experimentally on classification and regression benchmark data sets. 

\section{Problem Formulation}
We  focus on the widely studied supervised MCL setting where we let $\mR$ denote the set of real numbers and use boldface to denote vectors and matrices. We  use $\|.\|$ to denote the Euclidean norm for vectors and the Frobenius norm for matrices. The lifetime of the model is given by $[0,\Gamma]: \Gamma \in \mR$, where  $\Gamma$ is the maximum lifetime of the model. We  let $p(\cT)$ be the distribution over all the tasks in the interval $[0,\Gamma]$. Based on underlying processes that generate the tasks, the task arrival can be continuous time~(CT) or discrete time~(DT). For example, consider the system identification problem in processes modeled by  ordinary differential equations~(ODEs) or partial differential equations~(PDE) where the tasks represented by the states of the process $x(t)$ are generated in CT through the ODE. On the other hand, in the typical supervised learning setting, consider an image classification problem where each task comprises a set of images sampled from a discrete process and the tasks arrive in DT.  As in many previous MCL works, we  focus on DT MCL. However,  we  develop our theory for the CT MCL setting first because a CT MCL approach is broadly applicable to many domains. In Section 3.3 we provide a DPMCL approach for DT MCL setting by discretizing our theory.

A task $\cT(t)$ is a tuple of input-output pairs~$\{\cX(t), \cY(t)\}$ provided in the interval  $[t, t + \Delta t] \forall t \in [0,\Gamma), \Delta t \in \mR.$ We denote $(\vect{x}(t), \vect{y}(t)) \in \{\cX(t), \cY(t)\}.$  We define a parametric model $g(.)$ with parameters  $\hat{\vect{\theta}}$ such that $\hat{\vect{y}}(t) = g(\vect{x}(t); \hat{\vect{\theta}}(t)).$ Although we will use neural networks, any parametric model can be utilized with our framework. The catastrophic forgetting cost measures the error of the model on all the previous tasks (typically known as the meta learning phase \citep{finn2017model}); the generalization cost measures the error of the model on the new task (typically known as the meta testing phase \citep{finn2017model}). The goal in MCL is to minimize both the catastrophic forgetting cost and generalization cost for every $t \in [0,\Gamma)$.  Let us split the interval $ [0,\Gamma)$ as $[0,t] \cup (t, \Gamma),$ where the intervals $[0,t]$ and $(t, \Gamma)$ comprise  previous tasks and new task (i.e., the collection of all the task that can be observed in the interval, $(t, \Gamma)$), respectively. To take all the previous tasks into account, we define the instantaneous catastrophic forgetting cost $J(t; \hat{\vect{\theta} }(t))$ to be the integral of the loss function~$\ell(\tau)$ at any $t \in [0,\Gamma)$ as 
\begin{equation}
	\begin{aligned}
		J(t; \hat{\vect{\theta} }(t)) = \int_{\tau = 0}^{t} \gamma(\tau) \ell(\tau) d\tau,
	\end{aligned}
	\label{eq_cost_1}
\end{equation}
where $\ell(\tau)$ is computed on task $\cT(\tau)$ with $\gamma(\tau)$ being a parameter describing the contribution of this task to the integral. The value of $\gamma(\tau)$ is critical for the integral to be bounded~(details are provided in Lemma 1). Given a new task, the goal is to perform well on the new task as well as maintain the performance on the previous tasks. To this end, we  write $V(t;\hat{\vect{\theta}}(t)),$ as the cumulative cost~(combination of catastrophic cost and generalization cost) that is integrated over $[t, \Gamma).$ We therefore seek to minimize $V(t;\hat{\vect{\theta}}(t))$ and obtain the optimal value, $V^{*}(t)$, by solving the problem
\begin{equation}
	\begin{aligned}
		V^{*}(t)= min_{\hat{\vect{\theta}}(\tau) \in \Omega: t \leq \tau \leq \Gamma } \int_{\tau=t}^{\Gamma} J(\tau; \hat{\vect{\theta}}(\tau))d~\tau,
	\end{aligned}
	\label{eq_cost_2}
\end{equation}
where $\Omega$ is the compact set that implies that the parameters are initialized appropriately. Note from Eq.\ \eqref{eq_cost_2} that $V^{*}(t)$ is the optimal cost value over the complete lifetime of the model $[0, \Gamma)$. Since we have only the data corresponding to all the tasks in the interval $[0,t]$, solving Eq.\ \eqref{eq_cost_2} in its current form is intractable. To circumvent this issue, we  take a dynamic programming view of the MCL problem. We introduce a new theoretical framework where we model the learning process as a dynamical system. Furthermore, we derive conditions under which the learning process is stable and optimal using tools from the optimal control literature~\citep{lewis2012optimal}. 
\section{Theoretical Meta Continual Learning Framework}
We will recast  the problem defined in Eq.~\eqref{eq_cost_2} using ideas from dynamic programming, specifically Bellman's principle of optimality \citep{lewis2012optimal}. We treat the MCL problem as a dynamical system and describe the system using the following PDE:
\begin{equation}
	\begin{aligned}
		- \frac{\partial V^{*}(t)}{\partial t } & =  min_{\hat{\vect{\theta}}(t) \in \Omega}  \big[ J(t; \hat{\vect{\theta}}(t)) + J_{N}(t; \hat{\vect{\theta}}(t))   + \big(  V_{\hat{\vect{\theta}}(t)}^{*} \big)^{T} \Delta \hat{\vect{\theta}} \big] + \big( V_{\vect{x}(t)}^{*}  \big)^{T} \Delta \vect{x}(t),& 
	\end{aligned}
	\label{eq_M_PDE}
\end{equation}
where $V^{*}(t)$ describes the optimal cost (the left-hand side of Eq. \eqref{eq_cost_2}) and $(.)^{T}$ refers to the transpose operator. The notation $A_{(.)}$ denotes the partial derivative of $A$ with respect to $(.),$ for instance, $V_{\hat{\vect{\theta}}(t)}^{*} =  \frac{\partial V^{*}(t;\hat{\vect{\theta}}(t))}{\partial  \hat{\vect{\theta}}(t) }.$ The solution to the MCL problem is the parameter $\hat{\vect{\theta}}$ that minimizes the right-hand side of Eq.~\eqref{eq_M_PDE}. This involves minimizing the impact of introducing a new task on the optimal cost. The impact is quantified by the four terms in the right-hand side of Eq.~\eqref{eq_M_PDE}: the cost contribution from all the previous tasks~$J(t; \hat{\vect{\theta}}(t));$ the cost due to the new task~$J_{N}(t; \hat{\vect{\theta}}(t));$   the change in the optimal cost due to the change in the parameters $\big(V_{\hat{\vect{\theta}}(t)}^{*} \big)^{T} \Delta \hat{\vect{\theta}}$; and the change in the optimal cost due to change in the input (introduction of new task)~$ \big( V_{\vect{x}(t)}^{*}  \big)^{T} \Delta \vect{x}(t).$ Note that since  $\vect{y}(t)$ is a function of $\vect{x}(t)$, the changes in the optimal cost due to $\vect{y}(t)$ are captured by $\big( V_{\vect{x}(t)}^{*}  \big)^{T} \Delta \vect{x}(t)$. \textit{The full derivation for Eq.~\eqref{eq_M_PDE} from  Eq.\ \eqref{eq_cost_2} is provided in Appendix A.1.} Eq.~\eqref{eq_M_PDE} is also known as the Hamilton-Jacobi Bellman equation in optimal control \citep{lewis2012optimal} with the key difference that there is an extra term to quantify the changes due to the new task.

\subsection{Analysis}
Our formalism has two critical elements:  $\gamma(t)$, which quantifies the contribution of each task to catastrophic forgetting cost, and  the impact of the change in the input data distribution~$\Delta \vect{x}$ on learning, specifically while adapting to new tasks. We will analyze them next.

\textbf{Impact of $\gamma(t)$ on catastrophic forgetting cost:} In Eq. \eqref{eq_cost_1}, the integral is indefinite if $\Gamma \rightarrow \infty.$ If this indefinite integral does not have a convergence point, our MCL problem cannot be solved. The existence of the convergence point depends directly on the contribution of each task determined through $\gamma(t)$. In the next lemma and corollary, we will impose conditions on $\gamma(t)$ under which the cost $J(t;\hat{\vect{\theta} }(t))$ has a converging point.
\begin{lemma}
	Consider $t \in [0, \Gamma): \Gamma \rightarrow \infty$ and $J(t;\hat{\vect{\theta} }(t)) = \int_{\tau = 0}^{t}  \gamma(\tau) \ell(\tau) d\tau,$ and let $\epsilon \leq \ell(\tau) \leq L, \forall \epsilon >0$ and let $\ell$ be a continuous $\forall \tau \in [0, t]$ and bounded function. Let $\gamma(t)$ be a monotonically decreasing sequence such that $\gamma(t) \rightarrow 0,$ as $t \rightarrow \infty$  and $\int_{\tau = 0}^{\infty} \gamma(\tau) < M, M \in \mR$ Under these assumptions, $J(t;\hat{\vect{\theta} }(t))$ is convergent. \textit{Proof: See Appendix A.2}.
\end{lemma}
The following corollary is immediate.
\begin{corr}
	Let $J(t;\hat{\vect{\theta} }(t)) = \int_{\tau = 0}^{t}  \gamma(\tau) \ell(\tau) d\tau$,  and let  $\ell$  be continuous and $L \geq \ell(\tau) \geq \epsilon, \forall \tau, \epsilon >0$. Consider two cases for $\gamma(t)$ as $\gamma(t)>0 : \gamma(t) = c \text{ or } lim_{t\rightarrow \infty} \gamma(t)=\infty$. Then $J(t;\hat{\vect{\theta} }(t))$ is divergent. \textit{Proof:  See Appendix A.2.}
\end{corr}

In Lemma 1 and Corollary 1, we  consider only the scenario in which each task contributes nonzero cost to the integral. Corollary~1 implies that no methodology can perform perfectly on all the previous tasks, as  has been observed empirically~\citep{lin1992self}.  By choosing $\gamma(t),$ however, we can control the catastrophic forgetting by choosing which tasks to forget. In a typical setting larger weights are given to the recently observed tasks, and smaller weights are given to the older tasks. However, any choice of  $\gamma(t)$ that will keep $J(t;\hat{\vect{\theta} }(t))$ bounded is reasonable. 

\textbf{Impact of $\Delta \vect{x}(t)$ on learning:} We first present the following theorem.
\begin{thm}
	Let $V(t; \hat{\vect{\theta}}(t))= \int_{\tau = t}^{\Gamma}  J(\tau; \hat{\vect{\theta}}(\tau) )d\tau$, and let  Lemma 1 hold. Let the first derivative of $V(t; \hat{\vect{\theta}}(t))$  be $- \big[ J(t;\hat{\vect{\theta}}(t)) )  +  \big( V_{\hat{\vect{\theta}}(t)} \big)^{T} \Delta \hat{\vect{\theta}}(t)  + \big( V_{x} \big)^{T} \Delta x(t)\big].$ Let there be a compact set $\Omega$  such that $\hat{\vect{\theta}}(t) \in \Omega.$ Consider the assumptions $ ||J_{\hat{\vect{\theta}}(t)}||, >0, ||J_{x}|| \|\Delta x(t)\|<1, \|J_{x}\|>0$ and  $  \frac{\partial}{\partial (\vect{\theta}(t) ) } \int_{\tau=t}^{c} J(\tau,  \hat{\vect{\theta}}(\tau) )d~\tau~=~J(t,\hat{\vect{\theta}}(t)), c \in [t,t+\Delta t], \forall t \in [0, \Gamma]$  Let the update for the model be provided as $ \alpha(t)V_{\hat{\vect{\theta}}(t)}$ with $\alpha(t)>0$ being the learning rate.  Choose $\alpha(t)= \frac{1 - ||J_{x}|| \|\Delta x(t)\| }{\beta \| J_{\hat{\vect{\theta}}(t)}\|}$.
	The first derivative of $V(t;  \hat{\vect{\theta}}(t))$ is negative semi-definite and $V(t;  \hat{\vect{\theta}}(t))$ is ultimately bounded with the bound on $J(t;\hat{\vect{\theta}}(t))$ given as $J(t;\hat{\vect{\theta}}(t)) \leq \beta,$   where $\beta$ is a user defined threshold on the cost.  \textit{Proof: See Appendix A}.
\end{thm}
In Theorem~1, there are three main assumptions. First is a consequence of Lemma~1 where the contributions of each task to the cost must be chosen such that the cost is bounded and convergent. Second is the assumption of a compact set $\Omega.$  This assumption implies that if a weight value initialized from within the compact set, there will be a convergence to local minima. Third is the assumption that $\|J_{\hat{\vect{\theta}}(t)}\|>0 $ and $\|J_{x}\|>0$ are important to the proof and reflect through the choice of the learning rate, $\alpha(t)= \frac{1 - ||J_{x}|| \|\Delta x(t)\| }{\beta \| J_{\hat{\vect{\theta}}(t)}\|}.$ The condition $||J_{\hat{\vect{\theta}}(t)}|| = 0$  is well known in the literature as the vanishing gradient problem~\citep{pascanu2013difficulty}. On the other hand, intuitively, if $\|J_{x}\| = 0$, then the value of the cost $J$ will not change due to change in the input, and the learning process will stagnate.  Nevertheless, a large change in the input data distribution presents issues in the learning process.  Note that for Theorem 1 to hold, $\alpha(t)>0$, therefore, $\|J_{x}\| \|\Delta x(t)\|<1$.  Let $ \|\Delta \vect{x}(t)\| \leq b_{x},$ where $b_{x}$ is the upper bound on the change in the input data distribution. If $b_{x}$ is large~(going from predicting on images to understanding texts), the condition $\|J_{x}\| \|\Delta x(t)\|<1$ will be violated, and our approach will be unstable.

We can, however, adapt our model to the change in the input data-distribution $\Delta x(t)$ exactly if  we can explicitly track the change in the input. This type of adaptation can be  done easily when the process generating $\vect{x}(t)$ can be described by using an ODE or PDE. In traditional supervised learning settings, however, such a description is not possible. The issue highlights the need for strong representation learning methodologies where a good representation over all the tasks can  minimize the impact of changes in $\Delta x(t)$ on the performance of the MCL problem~\citep{javed2019meta,beaulieu2020learning}. Currently, in the literature, it is common to control the magnitude of $\Delta x(t)$ through normalization procedures under the assumption that all tasks are sampled from the same distribution. Therefore, for all practical purposes, we can choose $\alpha(t) \leq (\beta~||J_{\hat{\vect{\theta}}(t)}||)^{-1}$.  

\textbf{Connection to MAML, OML, and their variants:}
The optimization problem in MAML \citep{finn2017model} and OML \citep{finn2019online} can be obtained from Eq. \eqref{eq_M_PDE} by setting the third and the fourth terms to zero, which provides $
    - \frac{\partial V^{*}(t)}{\partial t } =  min_{\hat{\vect{\theta}}(t) \in \Omega} [ J(t; \hat{\vect{\theta}}(t)) + J_{N}(t; \hat{\vect{\theta}}(t)].
	\label{eq_MAML}$
MAML and OML have two loops. In the meta training step  the goal is to learn a prior on all the previous tasks by performing multiple updates with data from the previous task.  This can be done by optimizing the first term on the right-hand side in the equation above. In the meta testing phase, the goal is to generalize to new tasks, which is akin to optimizing the second term by using repeated updates. Furthermore, in MAML, the optimization in the meta testing phase is performed with the second-order derivatives. With the choice of different architectures for the neural network, all of the approaches that build on MAML and OML such as \citep{javed2019meta} and \citep{beaulieu2020learning} can be directly derived. All these methods do not adopt the PDE formalism; the third and fourth terms in Eq. \eqref{eq_M_PDE} are not explicitly included in MAML and OML.  On the other hand, the works in \citep{javed2019meta} and \citep{beaulieu2020learning} use a well-learned representation to implicitly minimize the impact of the last two terms~(impact of change in input distribution). In our DPMCL approach, we  explicitly consider these extra terms from Eq.~\eqref{eq_M_PDE} in the design of the weight update rule, providing us with a much more methodical process of addressing the impact of these terms and, by extension, the impact of key challenges in the MCL setting.

\subsection{Dynamic programming-based meta continual learning (DPMCL)}
As a consequence of Theorem 1, the update for the parameters is provided by $\alpha(t)  V_{\hat{\vect{\theta}}(t))}$. Since $V(t;\hat{\vect{\theta}}(t))$ is not completely known, we  have to approximate this gradient. To derive this approximation, we first rewrite Eq. \eqref{eq_M_PDE} as $ - \frac{\partial V^{*}(t; \hat{\vect{\theta}}(t))}{\partial t } = min_{\hat{\vect{\theta}}(t) \in \Omega}  \big[H(t;  \hat{\vect{\theta}}(t))\big],$ which provides the optimization problem as $\hat{\vect{\theta}}^{*}(t) =  \argmin_{\hat{\vect{\theta}}(t) \in \Omega}  \big[ H(t;  \hat{\vect{\theta}}(t))) \big],$ where $H(t;\hat{\vect{\theta}}(t)))~=~J(t; \hat{\vect{\theta}}(t)) + J_{N}(t; \hat{\vect{\theta}}(t)) + \big(  V^{*}_{\hat{\vect{\theta}}(t)} \big)^{T} \Delta \hat{\vect{\theta}} + \big( V^{*}_{\vect{x}(t)}  \big)^{T} \Delta \vect{x}(t)$  is the CT Hamiltonian. The solution for this optimization problem (the updates for the parameters in the network) is provided when the derivative of the Hamiltonian is set to zero.  First, we  discretize the MCL problem setting. Let $k$ be the discrete sampling instant such that $t = k \Delta t,$ where $\Delta t$ is the sampling interval.  Let a task $\cT_{k}$  at instant $k$ be sampled from $p(\cT).$ Let $\cT_{k} =(\cX_{k},\cY_{k})$ be a tuple, where $\cX_{k} \in \mR^{n \times p}$ denotes the input data and  $\cY_{k} \in \mR^{n \times 1}$ denotes the target labels (output). Let $n$ be the number of samples and  $p$ be the number of dimensions. Let the parametric model be given as   $\hat{\vect{y}} = g( h( \vect{x} ; \hat{\vect{\theta}}_{1}); \hat{\vect{\theta}}_2),$ where the inner map $ h(.)$ is treated as a representation learning network and $g(.)$ is the prediction network. Let $\hat{\vect{\theta}} = [\hat{\vect{\theta}}_{1} \quad \hat{\vect{\theta}}_2]$ be the learnable model parameters and the weight updates be given as
\begin{equation}
	\begin{aligned}
		\hat{\vect{\theta}}(k+1) &=&\hat{\vect{\theta}}(k) - \alpha(k) \frac{\partial}{\partial \hat{\vect{\theta}}(k) } \big[ J_{N}(k) +  J_{P}(k) + \big(J_{PN}(k) - J_{PN}(k; \hat{\vect{\theta}}(k+\zeta))\big) \big]
	\end{aligned}
	\label{eq_update}
\end{equation}
Our update rule has three terms~(the terms inside the bracket). The first term depends on $J_{N}$, which is calculated on the new task; the second term depends on $J_{P}$ and is calculated on all the previous tasks; the third term comprises  $J_{PN}$ and is evaluated on a combination of the previous tasks and the new task. The first term minimizes the generalization cost. Together, the second and the third terms minimize the catastrophic forgetting cost. The second term minimizes the cost evaluated on just the previous tasks, whereas the third term reduces the impact of change in input and the weights introduced by the presence of the new task (product of the third and the fourth terms in Eq. \eqref{eq_M_PDE}). If we set the third term to zero, we can achieve the update rule for MAML and OML (the first-order approximation). In Eq.~\eqref{eq_update} $\alpha(k) \leq \frac{1}{\beta \|J_{\hat{\vect{\theta}(k))}}\|^2+\epsilon},$ where $\beta$ is a user-defined parameter and $\epsilon>0$ is a small value to ensure that the denominator does not go to zero.~\textit{The derivation of the update rule and the discretization are presented in Appendix~A.3}.

Equipped with the gradient updates, we now describe the DPMCL algorithm. We define a new task sample, $ \cD_N(k) = \{ \mathcal{X}_{k}, \mathcal{Y}_{k} \},$ and a task memory (samples from all the previous tasks) $\cD_{P}(k) \subset \cup_{\tau = 0}^{k-1}\cT_{\tau}$. We can approximate the required terms in our update rule Eq.~\ref{eq_update} using samples (batches) from $\cD_{P}(k)$ and $\cD_N(k).$  The overall algorithm consists of two steps: generalization  and  catastrophic forgetting~(see Algorithm 1 in Appendix B.4). DPMCL comprises representation and prediction neural networks parameterized by $\hat{\vect{\theta}}_{1}$ and $\hat{\vect{\theta}}_2$, respectively. For each batch $b_{N} \in \cD_N(k)$, DPMCL alternatively performs generalization and catastrophic forgetting cost updates $\kappa$ times. The generalization cost update consists of computing the cost $J_N$ and using that to update $\hat{\vect{\theta}}_{1}$ and $\hat{\vect{\theta}}_2$; the catastrophic forgetting cost update comprises the following steps. First we create a batch that combines the new task data with samples from the previous tasks $b_{PN} = b_{P} \cup b_{N}(k)$, where $b_{P} \in \cD_{P}(k)$. Second, to approximate the term $\big( J_{PN}(k; \hat{\vect{\theta}}(k+\zeta))),$,  we  copy $\hat{\vect{\theta}}_{2}$~(prediction network) into a temporary network parameterized by $\hat{\vect{\theta}}_{B}$. We  then perform $\zeta$ updates on $\hat{\vect{\theta}}_{B}$ while keeping $\hat{\vect{\theta}}_{1}$ fixed. Third, using $\hat{\vect{\theta}}_{B}(k+\zeta),$ we compute $J_{PN}(k;\hat{\vect{\theta}}_{B}(k+\zeta))$ and update $\hat{\vect{\theta}}_{1}, \hat{\vect{\theta}}_2$ with $J_{P}(k)+( J_{PN}(k) - J_{PN}(k;\hat{\vect{\theta}}_{B}(k+\zeta)) )$. The inner loop with $\zeta$ is purely for the purpose of approximating the optimal cost.

The rationale behind repeated updates to approximate $ J_{PN}(k; \hat{\vect{\theta}}(k+\zeta)),$ is as follows. At every instant $k,$ $J_{PN}(k)$~(the cost on all the previous tasks and the new task) is the boundary value for the optimal cost as the optimal cost can only be less than $J_{PN}(k)$. Therefore, if we start from $J_{PN}(k)$ and execute a Markov chain for $\zeta$ steps, the end point of this Markov chain is the optimal value of  $J_{PN}(k)$ at instant $k,$ provided $\zeta$ is large enough. Furthermore, the difference between the cost at the starting point and the end point of this chain provides us with a value that should be minimized such that the cost $J_{PN}(k)$ will reach the optimal value for the MCL problem at instant $k$. To execute this Markov chain, we perform repeated updates on a copy of $\hat{\vect{\theta}}_2$~(prediction network) denoted as $\hat{\vect{\theta}}_{B}.$ The goal of the representation network is to learn a robust representation across all tasks. If the representation network is updated multiple times with respect to each batch of data, it might get biased toward the data  present in the batch. Therefore, we keep the representation network fixed and update only a copy of $\hat{\vect{\theta}}_2$. We repeat this alternative update process for each data batch in the new task. Once all the data from the new task is exhausted, we  move to the next task.
\subsection{Related work}
Existing MCL methods can be grouped into three classes: (1) dynamic architectures and flexible knowledge representation \citep{sutton1990integrated, rusu2016progressive, yoon2017lifelong}; (2) regularization approaches, (\citep{kirkpatrick2017overcoming, zenke2017continual, aljundi2018memory}) and (3) memory/experience replay, \citep{lin1992self, chaudhry2019continual, lopez2017gradient}. Flexible knowledge representations maintain a state of the whole dataset with the advent of a new tasks that require computationally expensive mechanisms~\citep{yoon2017lifelong}. Regularization approaches \citep{sutton1990integrated,  lin1992self, chaudhry2019continual, lopez2017gradient} attempt to minimize the impact of new tasks~(changes in the input data-distribution) on the parameters of the model involving a significant trial-and-error process. Memory/experience replay-driven approaches~\citep{lin1992self}, \citep{chaudhry2019continual, lopez2017gradient} can address catastrophic forgetting but do not generalize well to new tasks

More recently, MCL has been investigated in \citet{finn2017model} and \citet{finn2019online}.  In \citet{finn2017model}, the authors presented a method where an additional term is introduced into the cost function (the gradient of the cost function with respect to the previous tasks). However, this method requires all the data to be known prior to the start of the learning procedure.  To obviate this constraint, an online meta-learning approach was introduced in \citet{finn2017model}. The approach does not explicitly minimize catastrophic forgetting but focuses on fast online learning. In contrast with \citet{finn2017model}, our method can learn sequentially as the new tasks are observed. Although sequential learning was possible in \citet{finn2019online}, the work highlighted the trade-off between memory requirements and catastrophic forgetting, which must to be addressed by learning a representation over all the tasks as in \citet{javed2019meta, beaulieu2020learning}. Similar to \citet{javed2019meta, beaulieu2020learning}, our method allows a representation to be learned over the distribution of all the tasks $p(\cT)$. However, both the representation and the training model are learned sequentially in DPCML, and we do not observe a pretraining step.

Our approach is the first comprehensive theoretical framework based on dynamic programming that is model agnostic and can adapted to different MCL setting  in both CT and DT. Although theoretical underpinnings were provided in  \citet{finn2019online} and \citet{flennerhag2019metalearning}, the focus was to provide structure for parameter updates but not attempt to holistically model the overall learning dynamics as is done in our theoretical framework. The key ideas in this paper have been adapted from dynamic programming and optimal control theory; additional details can be found in \citet{5227780}.

\section{Experiments}
 We use four continual learning data sets: incremental sine wave (regression on 50 tasks \textit{(SINE)}); split-Omniglot (classification on 50 tasks, \textit{(OMNI)}); continuous MNIST (classification on 10 tasks and \textit{(MNIST)}); and CIFAR10 (classification on 10 tasks, \textit{(CIFAR10)}). All these data sets have been used in \citep{finn2019online, finn2017model}. We compare DPMCL with Naive (training is always performed on the new task without any explicit catastrophic forgetting minimization); Experience-Replay (ER) \citep{lin1992self} (training is performed by sampling batches of data from all the tasks (previous and the new)); online-meta learning (OML)  \citep{finn2019online}, online meta-continual learning (CML) \citep{javed2019meta}, neuro-modulated meta learning (ANML) \citep{beaulieu2020learning}. To keep consistency with our computing environment and the task structure, we implement a sequential online version \citep{finn2019online} (where each task is exposed to the model sequentially) of all these algorithms in our environment. For any particular data set, we set the same model hyper-parameters~(number of hidden layers, number of hidden layer units, activation functions, learning rate, etc.) across all implementations. For fair comparisons, for any given task we also fix the total number of gradient updates a method can perform~(See Appendix~B.1,2,3 for details on data-set and hyper-parameters).

 For each task, we split the given data into training (60\%), validation (20\%), and testing (20\%) data sets. Methods such as ANML, OML, and CML follow the two loop training strategy from OML: the inner loop and the outer loop. The training data for each task is used for the inner loop, whereas the validation data is used in the outer loop. The testing data is used to report accuracy metrics. We measure generalization and catastrophic forgetting through cumulative error (CME) given by the average error on all the previous tasks and the new task error (NTE) given by the average error on the new task, respectively. For regression problems, they are computed from mean squared error; for classification problems, given as $(1- \frac{Acc}{100}),$ where $Acc$ refers to the classification accuracy. For cost function, we use the mean squared error for regression and categorical cross-entropy for classification. We use a total of 50 runs~(repetition) with different random seeds and report the mean $\mu$ and standard error of the mean ($\sigma_{error} = \sigma/\sqrt{50},$ where $\sigma$ is the standard deviation with 50 being the number of repetition). We  report only the $\sigma_{error}$ when it is greater than $10^{-3}$; otherwise, we indicate a $0$~(See Appendix~B.4,5 for implementations details).
 
\subsection{Results}
\begin{figure}[!tb]
		\centering
		\includegraphics[ width = 0.9\textwidth]{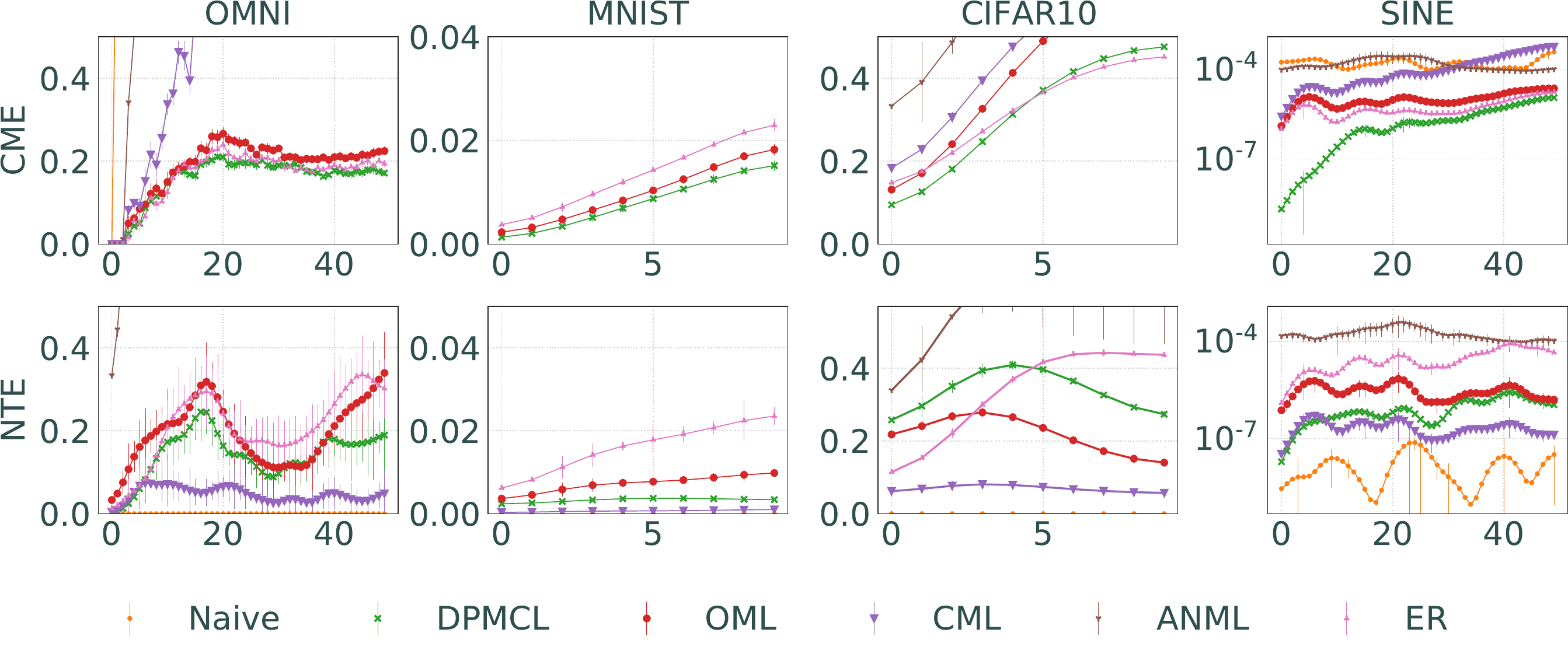}
	\caption{Top row: Cumulative error~(CME) trends with respect to tasks; Bottom row: new task error~(NTE) trends with respect to tasks. The error bars describe the area between $\mu + \sigma_{error}$ and $\mu - \sigma_{error}$ over  50 repetitions. 
	Gaussian smoothing filter with standard deviation of 2 is applied on each trajectory. ~\label{fig:result}}
\end{figure}

We first analyze the \texttt{CME} and \texttt{NTE} as each task is incrementally shown to the model. We record the \texttt{CME} and \texttt{NTE} on the testing data~(averaged over 50 repetitions) at each instant when a task is observed. The results are shown in Fig. \ref{fig:result}. Unlike the other methods,  DPMCL achieves low error with respect to \texttt{CME} and \texttt{NTE}.  ANML and CML perform poorly on \texttt{CME} because of the lack of a learned representation~(as we do not have a pretraining phase to learn an encoder). The performance of DPMCL is better than OML and ER in all data sets except CIFAR-10, where ER is comparable to DPMCL. The poor performance of Naive is expected because it is trained only on the new task data and thus incurs catastrophic forgetting. DPMCL generalizes well to new tasks, and consequently the performance of DPMCL is better than OML, ER, and ANML in all the data sets except CIFAR10. As expected, Naive has the lowest \texttt{NTE}. In the absence of a well-learned representation, CML exhibits behavior similar to Naive's and is able to quickly generalize to new task. For CIFAR10, OML achieves \texttt{NTE} that is lower than that of DPMCL. ER struggles to generalize to a new task; however, the performance is better than ANML's, which exhibits the poorest performance due to the absence of a well-learned representation.

The \texttt{CME} and \texttt{NTE} values for all the data sets and methods are summarized in Table 1 in Appendix B.6. DPMCL achieves \texttt{CME} $[\mu (\sigma_{err})]$ of $10^{-5} (0)$ for SINE, $0.171(0.007)$ for OMNI, $0.020(0.001)$ for MNIST, and $0.496(0.003)$ for CIFAR10. We note that the \texttt{CME} for DPMCL are the best among all the methods and all the data sets except OML, ER for SINE, and ER for CIFAR10. In SINE, the \texttt{CME} of DPMCL are comparable to OML and ER. In CIFAR10, ER demonstrates a $3 \%$ improvement in accuracy~($(0.496-0.464)\times100$). On the \texttt{NTE}~$[\mu~(\sigma_{err})]$ scale, DPMCL achieves $10^{-7} (0)$ for SINE, $0.283(0.091)$ for OMNI, $0.003(0)$ for MNIST and $0.231(0.008)$ for  CIFAR10. On the \texttt{NTE} scale, the best-performing methods are CML and Naive; this behavior can be observed in the trends from Fig. \ref{fig:result}.  DPMCL is better than all the other methods in all the data sets except CIFAR 10, where OML is better~(observed earlier in Fig. \ref{fig:result}) by 10\% ~($(0.231-0.108)\times100$). However,  this 10\% improvement for OML comes at the expense of 18 \%~($(0.676-0.496)\times100$) drop in performance on the \texttt{CME} scale. Similarly, although ER exhibits a 3 \% improvement on \texttt{CME} scale, DPMCL outperforms ER on the \texttt{NTE} scale by a 22.7 \%~($(0.458-0.231)\times100$) improvement. The results show that DPMCL achieves a balance between \texttt{CME} and \texttt{NTE}, thus performing better than or comparable to the state of the art in the MCL setting.

\begin{figure}
		\centering
		\includegraphics[ height = 1.3in, width =0.85\columnwidth]{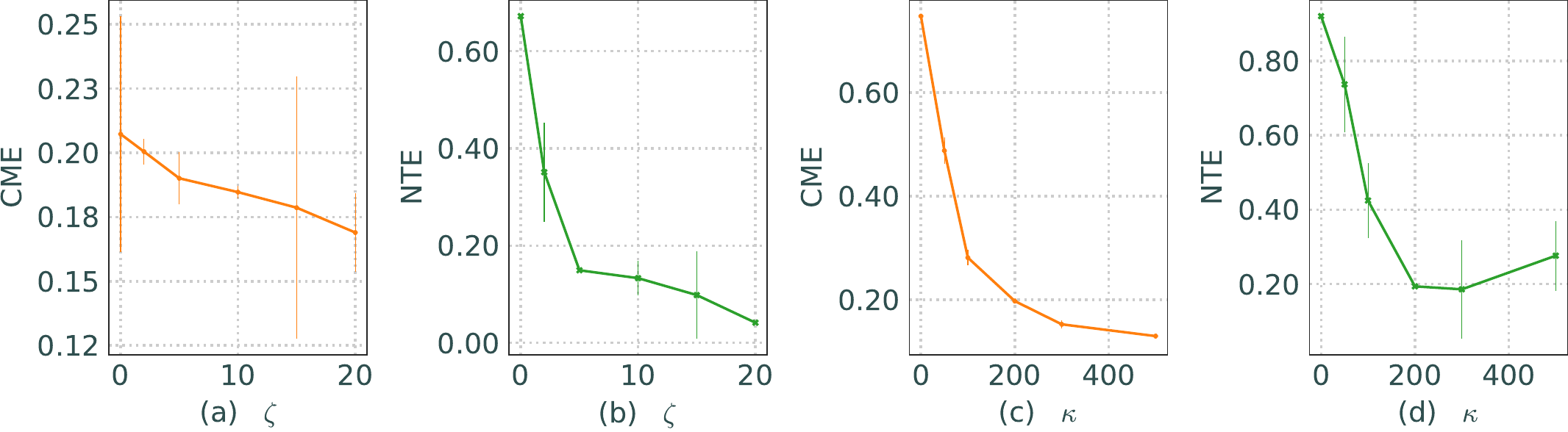}
	\caption{Cumulative error (CME) and new task error~(NTE) trends with respect to (a,b) $\zeta$ with $\kappa = 200$ and (c, d) $\kappa$ with $\zeta = 2.$ \label{fig:hyper}. For each value of $\kappa$ or $\zeta,$ we learn a total of 50 tasks incrementally. We perform 50 repetitions of this learning. Next, we calculate the $\mu$ and $\sigma_{err}$ value of cumulative error and new task error over these 50 runs and record them. After we have computed these values for each value of $\kappa$ or $\zeta,$ we plot the mean values with error bars from $\sigma_{err}$ as a function of $\zeta$ or $\kappa.$ We apply a Gaussian smoothing filter with a standard deviation of 2.}
\end{figure}

The improved performance of DPMCL is because of the third term in Eq. \eqref{eq_update}. Therefore, we analyze the impact of the term and show that without the third term, DPMCL suffers from poor generalization and catastrophic large forgetting. In Fig. \ref{fig:hyper}, we plot the variation in \texttt{CME} and \texttt{NTE} for the $50^{th}$ task in the OMNI data set with respect to hyperparameter $\zeta.$ In DPMCL, the value of $\zeta$ controls the magnitude of the third term in the update rule, namely, $\big(J_{PN}(k) - J_{PN}(k; \hat{\vect{\theta}}(k+\zeta)))\big),$ where $J_{PN}(k; \hat{\vect{\theta}}(k+\zeta)))$ is an approximation of the optimal cost. Therefore, the larger the value of $\zeta,$ the greater the value of $\big(J_{PN}(k) - J_{PN}(k; \hat{\vect{\theta}}(k+\zeta)))\big)$, and the third term is zero when $\zeta = 0. $ We observe from  Figs. \ref{fig:hyper}(a) and (b) that when the value of $\zeta$ is zero, DPMCL generalizes poorly to a new task~(low \texttt{NTE}) and incurs catastrophic forgetting~(low \texttt{CME}).  As the value of $\zeta$ is increased from zero, however, forgetting reduces~(decreasing \texttt{CME} in Fig. \ref{fig:hyper}(a)) and generalization improves~(decreasing \texttt{NTE} in Fig. \ref{fig:hyper}(b)). Furthermore, the best value of forgetting with generalization is achieved when $\zeta = 20.$ We know from our theoretical analysis that the larger the value of $\zeta,$ the closer $J_{PN}(k; \hat{\vect{\theta}}(k+\zeta))$ is to the optimal cost. Minimizing the difference $\big(J_{PN}(k) - J_{PN}(k; \hat{\vect{\theta}}(k+\zeta)))\big)$ would push the model toward optimal generalization and catastrophic forgetting. This behavior is observed in Fig. \ref{fig:hyper}(a, b).

Next, we analyzed the impact of $\kappa,$ the parameter controlling the total number of alternative updates. From our theory, we know that an appropriate number of $\kappa$ allows DPMCL to balance forgetting and generalization. We observe this behavior from  Figs. \ref{fig:hyper}(c) and (d).  Note that with an increase in $\kappa$ from zero, forgetting keeps on reducing~(decreasing \texttt{CME} on Fig. \ref{fig:hyper}(c).)  When $\kappa > 250, $ however, we observe that while forgetting does improve, generalization no longer improves~(Fig. \ref{fig:hyper}(d)). In fact, we observe an inflection point on \texttt{NTE}  between $\kappa = 200$ and $\kappa = 250.$ where a balance between \texttt{CME} and \texttt{NTE} is achieved. DPMCL is designed in such a way that with choice of $\kappa$, this balance point can be engineered. 

\section{Conclusions}
We introduced a dynamic-programming-based theoretical framework for meta continual learning.Within this framework, catastrophic forgetting and generalization, the two central challenges of the meta continual learning, can be studied and analyzed methodically. Furthermore, the framework also allowed us to provide theoretical justification for intuitive and empirically proven ideas about generalization and catastrophic forgetting. We then introduced DPMCL which was able to systematically model and compensate for the trade-off between the catastrophic forgetting and generalization. We also provided experimental results in a sequential learning setting that show that the framework is practical with comparable performance to state of the art in meta continual learning. 

In the future, we plan to extend this approach to reinforcement and unsupervised learning. Moreover, we plan to study different architectures such as convolutional neural networks and graph neural networks.

\subsubsection*{Author Contributions}
If you'd like to, you may include  a section for author contributions as is done in many journals. This is optional and at the discretion of the authors.
\subsubsection*{Acknowledgments}
Use unnumbered third level headings for the acknowledgments. All
acknowledgments, including those to funding agencies, go at the end of the paper.
\bibliographystyle{plain}

\appendix
\section{Appendix -Derivation and Proofs}
\label{proofs}
First we will derive our PDE.
\subsection{Derivation of Hamilton-Jacobi Bellman for the Meta Continual Learning Setting}
Let the optimal cost be given as 
\begin{equation}
	\begin{aligned}
		V^{*}(t;\hat{\vect{\theta}}(t))= min_{\hat{\vect{\theta}}(\tau) \in \Omega: t \leq \tau \leq \Gamma }  \bigg[\int_{\tau=t}^{\Gamma}  J(\tau; \hat{\vect{\theta}}(\tau))d~\tau \bigg].
	\end{aligned}
	\label{eq_a_1_1}
\end{equation}

We  split the interval $[t, \Gamma]$ as $[t, t+ \Delta t]$, and $[t+ \Delta t, \Gamma].$ With this split, we  rewrite the cost function as 
\begin{equation}
	\begin{aligned}
		V^{*}(t;\hat{\vect{\theta}}(t)) = min_{\hat{\vect{\theta}}(\tau) \in \Omega: t \leq \tau \leq\Gamma}  \bigg[ \int_{\tau=t}^{t + \Delta t}  J(\tau; \hat{\vect{\theta}}(\tau))  d~\tau 
		\\ + \int_{\tau = t + \Delta t}^{\Gamma}  J(\tau; \hat{\vect{\theta}}(\tau))d~\tau \bigg].
	\end{aligned}
	\label{eq_a_1_2}
\end{equation}
With $V(t;\hat{\vect{\theta}}(t) )=  \int_{\tau=t}^{\Gamma}  J(\tau; \hat{\vect{\theta}}(\tau))d~\tau,$ note that $\int_{\tau=t + \Delta t}^{\Gamma}  J(\tau; \hat{\vect{\theta}}(\tau))d~\tau$ is $V$ at $t + \Delta t$ and can be defined as $ V(t+ \Delta t; \hat{\vect{\theta}}(t+ \Delta t)) $,  which provides
\begin{equation}
	\begin{aligned}
		V^{*}(t;\hat{\vect{\theta}}(t)) = min_{\hat{\vect{\theta}}(\tau) \in \Omega: t \leq \tau \leq\Gamma}  \bigg[ \int_{\tau=t}^{t + \Delta t} J(\tau; \hat{\vect{\theta}}(\tau))  d~\tau 
		\\ +  V(t+ \Delta t; \hat{\vect{\theta}}(t+ \Delta t)) \bigg].
	\end{aligned}
	\label{eq_opt}
\end{equation}
Suppose now that all information for $\tau \geq t+ \Delta t$ is known, and also suppose that all optimal configurations of parameters are  known.  With this information, we  can narrow our search to just the optimal costs for the interval $[t, t+ \Delta t]$ and write
\begin{equation}
	\begin{aligned}
		V^{*}(t;\hat{\vect{\theta}}(t)) = min_{\hat{\vect{\theta}}(\tau) \in \Omega: t \leq \tau \leq t + \Delta t}  \bigg[ \int_{\tau=t}^{t + \Delta t}  J(\tau; \hat{\vect{\theta}}(\tau))  d~\tau
		\\ +  V^{*}(t+ \Delta t; \hat{\vect{\theta}}(t+ \Delta t))\bigg].
	\end{aligned}
	\label{eq_a_1_4}
\end{equation}
Now, the only parameters to be obtained are for the sequence $t \leq \tau \leq t + \Delta t.$ We   approximate the $V^{*}(t+ \Delta t; \hat{\vect{\theta}}(t+ \Delta t))$ using the information provided in the interval  $[t, t+ \Delta t].$ To do so, we  further simplify this framework by writing the first-order Taylor series expansion. However, $V^{*}(t+ \Delta t; \hat{\vect{\theta}}(t+ \Delta t))$ is a function of $\vect{y}(t),$ that is the model. Since $\vect{y}(t)$ is a function of $(t, \vect{x}(t), \hat{\vect{\theta}}(t)),$ all changes in  $\vect{y}(t),$ can be summarized through $(t, \vect{x}(t), \hat{\vect{\theta}}(t)).$
Therefore, we evaluate the Taylor series around $(t, \vect{x}(t), \hat{\vect{\theta}}(t)),$

\begin{equation}
	\begin{aligned}
		V^{*}(t+ \Delta t; \hat{\vect{\theta}}(t+ \Delta t)) &= V^{*}(t;\hat{\vect{\theta}}(t)) + \big(V_{t}^{*} \big)^{T} \Delta t & \\ &+ \big(  V_{\hat{\vect{\theta}}(t)}^{*} \big)^{T} \Delta \hat{\vect{\theta}} + \big( V_{\vect{x}(t)}^{*}  \big)^{T} \Delta \vect{x}(t),&
	\end{aligned}
	\label{eq_a_1_5}
\end{equation}
where we use the notation $V_{(.)}^{*}$ to denote the partial derivative with respect to $(.).$ For instance, 
$V_{\hat{\vect{\theta}}(t)}^{*} =  \frac{\partial V^{*} ((t;\hat{\vect{\theta}}(t))}{\partial  \hat{\vect{\theta}}(t) }.$ Substituting into the original equation, we have 
\begin{equation}
	\begin{aligned}
		V^{*}(t;\hat{\vect{\theta}}(t)) &= min_{\hat{\vect{\theta}}(\tau) \in \Omega: t \leq \tau \leq t + \Delta t}  \bigg[ \int_{\tau=t}^{t + \Delta t} J(\tau;\hat{\vect{\theta}}(\tau))  d~\tau   & \\ & + V^{*}(t;\hat{\vect{\theta}}(t))  + \big(V_{t}^{*}   \big)^{T} \Delta t + \big(  V_{\hat{\vect{\theta}}(t)}^{*} \big)^{T} \Delta \hat{\vect{\theta}}  & \\ & + \big( V_{\vect{x}(t)}^{*}  \big)^{T} \Delta \vect{x}(t)\bigg] & .
	\end{aligned}
	\label{eq_a_1_6}
\end{equation}
The terms $V^{*}(t;\hat{\vect{\theta}}(t)) + \big(V_{t}^{*}   \big)^{T} \Delta t$ can be brought outside the minimization because they are independent of $\tau$, the sequence being selected. Therefore,
\begin{equation}
	\begin{aligned}
		V^{*}(t;\hat{\vect{\theta}}(t)) &= min_{\hat{\vect{\theta}}(\tau) \in \Omega: t \leq \tau \leq t + \Delta t }  \bigg[ \int_{\tau=t}^{t + \Delta t} J(\tau; \hat{\vect{\theta}}(\tau))  d~\tau & \\ &+ \big(  V_{\hat{\vect{\theta}}(t)}^{*} \big)^{T} \Delta \hat{\vect{\theta}} + \big( V_{\vect{x}(t)}^{*}  \big)^{T} \Delta \vect{x}(t) \bigg] + V^{*}(t;\hat{\vect{\theta}}(t))  & \\ &  + \big(V_{t}^{*}   \big)^{T} \Delta t    & .
	\end{aligned}
	\label{eq_a_1_7}
\end{equation}
Upon cancellation of common terms we have 
\begin{equation}
	\begin{aligned}
		-  \big(V_{t}^{*}   \big)^{T} \Delta t & = min_{\hat{\vect{\theta}}(\tau) \in \Omega: t \leq \tau \leq t + \Delta t }  \bigg[ \int_{\tau=t}^{t + \Delta t}  J(\tau; \hat{\vect{\theta}}(\tau))  d~\tau   & \\ &  + \big(  V_{\hat{\vect{\theta}}(t)}^{*} \big)^{T} \Delta \hat{\vect{\theta}} + \big( V_{\vect{x}(t)}^{*}  \big)^{T} \Delta \vect{x}(t)\bigg]. &
	\end{aligned}
	\label{eq_a_1_8}
\end{equation}
Observe that $\int_{\tau=t}^{t + \Delta t}  J(\tau; \hat{\vect{\theta}}(\tau))  d~\tau  = J(t; \hat{\vect{\theta}}(t)) + \int_{\tau = t+}^{t+\Delta t} \gamma( \tau )\ell(\tau) d\tau,$ where $J(t; \hat{\vect{\theta}}(t))$ represents all the previous tasks and $\int_{\tau = t+}^{t+\Delta t} \gamma( \tau )\ell(\tau) d\tau$ represents the new task. Therefore, we get our final PDE as 
\begin{equation}
	\begin{aligned}
		-  \big(V_{t}^{*}   \big)^{T} \Delta t & = min_{\hat{\vect{\theta}}(\tau) \in \Omega: t \leq \tau \leq t + \Delta t }  \bigg[ J(\tau; \hat{\vect{\theta}}(\tau)) & \\ &  +  \int_{\tau = t+}^{t+\Delta t} \gamma( \tau )\ell(\tau) d\tau& \\ &  + \big(  V_{\hat{\vect{\theta}}(t)}^{*} \big)^{T} \Delta \hat{\vect{\theta}} + \big( V_{\vect{x}(t)}^{*}  \big)^{T} \Delta \vect{x}(t)\bigg]. &
	\end{aligned}
	\label{eq_a_1_9}
\end{equation}
Let us push $\Delta t \rightarrow 0$ and denote $lim_{\Delta t \rightarrow 0} \int_{\tau = t+}^{t+\Delta t} \gamma( \tau )\ell(\tau) d\tau$ as $J_{N}(\tau; \hat{\vect{\theta}}(\tau)).$ We get 
\begin{equation}
	\begin{aligned}
		-  \big(V_{t}^{*}   \big)^{T} \Delta t & = min_{\hat{\vect{\theta}}(t) \in \Omega}  \bigg[ J(t; \hat{\vect{\theta}}(t)) + J_{N}(t; \hat{\vect{\theta}}(t)) & \\ &  + \big(  V_{\hat{\vect{\theta}}(t)}^{*} \big)^{T} \Delta \hat{\vect{\theta}} + \big( V_{\vect{x}(t)}^{*}  \big)^{T} \Delta \vect{x}(t)\bigg]. &
	\end{aligned}
	\label{eq_a_1_10}
\end{equation}

This is the Hamilton-Jacobi Bellman equation that is specific to the meta continual learning problem.
\subsection{Proof of Lemma and Corollary 1}
\begin{proof}[Proof of Lemma 1]
	Consider $J(t;\hat{\vect{\theta} }(t)) = \int_{\tau = 0}^{t}  \gamma(\tau) \ell(\tau) d\tau.$  Consider $t_{1}, t_{2} \in [0, t): t_{2}>t_{1}$. By the ratio test $\lvert \frac{\gamma(t_{2})}{\gamma(t_{1})} \rvert <1$ and $\int_{\tau = 0}^{t} \gamma(\tau) d\tau$ is convergent. Invoking the condition $\ell(\tau) \leq L, \forall \tau$, we get  $J(t;\hat{\vect{\theta} }(t)) \leq \int_{\tau = 0}^{t}  \gamma(\tau) L d\tau.$ Since $\gamma(\tau)$ is convergent, by linearity $ \int_{\tau = 0}^{t}  \gamma(\tau) L d\tau.$ is convergent. Therefore, $J(t;\hat{\vect{\theta} }(t))$ is upper bounded by a sequence that is convergent. Thus $J(t;\hat{\vect{\theta} }(t))$ is convergent.
\end{proof}

\begin{proof}[Proof of Corollary 1]
	We know that the cost is given as $J(t;\hat{\vect{\theta} }(t)) = \int_{\tau = 0}^{t}  \gamma(\tau) \ell(\tau) d\tau$ . We consider two cases. In the first case $\gamma(t) = c,$, where $c$ is a constant. One can easily see that $J(t;\hat{\vect{\theta} }(t)) \geq \lim_{t \rightarrow \infty }\int_{t=0}^{\infty} c \epsilon dt,$ where $\lim_{t \rightarrow \infty }\int_{t=0}^{\infty} c \epsilon dt =  \lim_{t \rightarrow \infty } [c\epsilon t]_{0}^{\Gamma} = \infty.$ 
	
	In the second case, $J(t;\hat{\vect{\theta} }(t)) \geq \int_{t=0}^{\infty} \epsilon \gamma(t) dt, \forall \epsilon>0.$ Consider $lim_{t\rightarrow \infty} \gamma(t) = \infty$ and let $t_{1}, t_{2} \in [0, \infty): t_{2} > t_{1}.$ By the ratio test, $| \frac{\gamma(t_{2})}{\gamma(t_{1})} | >1$ and $ \int_{t=0}^{\infty} \gamma(t) dt$ is divergent. Therefore, $J(t;\hat{\vect{\theta} }(t))$ is lower bounded by a function that is divergent.
\end{proof}

\subsection{Proof of Theorem 1}
\begin{proof}[Proof of Theorem 1]
	
	To show that the cumulative cost will achieve a bound, we need only  to show that the first derivative of the cumulative cost is negative semi-definite and bounded. This idea stems from the principles of Lyapunov, which we discuss below.
	
	\noindent \textbf{Lyapunov Principles}
	The basic idea is to demonstrate stability of the system described by the PDE in the sense of Lyapunov.
	\begin{defn}[Definition of stability in the Lyapunov sense (\citep{lewis2012optimal})]
		Let $V(x, t)$ be a non-negative function with derivative $\dot{V}(x,t)$ along the system. The following is then true:
		\begin{enumerate}
			\item  If $V(x,t)$ is locally positive definite and $\dot{V}(x,t) \leq 0$ locally in $x$ and for all $t$, then the equilibrium point is locally stable (in the sense of Lyapunov).
			\item If $V(x,t)$ is locally positive definite and decresent and $\dot{V}(x,t) \leq 0$ locally in $x$ and for all $t$, then the equilibrium point  is uniformly locally stable (in the sense of Lyapunov).
			\item  If $V(x,t)$ is locally positive definite and decresent and $\dot{V}(x,t) < a$ in $x$ and for all $t$, then the equilibrium point  is Lyapunov stable, and the equillibrium point is ultimately bounded. Refer to  Definition \ref{UUB}.
			\item  If $V(x,t)$ is locally positive definite and decresent and $\dot{V}(x,t) < 0$ locally in $x$ and for all $t$, then the equilibrium point is locally asymptotically stable.
			\item  If $V(x,t)$ is locally positive definite and decresent and $\dot{V}(x,t) < 0$ in $x$ and for all $t$, then the equilibrium point  is globally asymptotically stable.
		\end{enumerate}
	\end{defn}
	In our analysis we seek to determine the behavior of cumulative cost with respect to change in the parameters (controllable by choice of the parameter update) and change in the input (uncontrollable). We assume that the changes due to input  and parameter update are both bounded, and we conclude that V is ultimately bounded and stable in the sense of Lyapunov. The idea of ultimately bounded is described by the next definition.
	
	\begin{defn}
		The solution of a differential equation $\dot{x} = f(t,x)$ is uniformly ultimately bounded with ultimate bound b if $b$ and $c$ and for every $0<a<c$, $\exists T =T(a,b) \geq 0$ such that $\|x(t0)\| \leq a \implies \|x(t)\|\leq b, \forall t \geq t0+T.$
		\label{UUB}
	\end{defn}
	The full proof is  as follows. Let the Lyapunov function be given as
	\begin{equation}
		\begin{aligned}
			V(t;\hat{\vect{\theta}}(t)) = \int_{\tau=t}^{\Gamma} J(\tau,  \hat{\vect{\theta}}(\tau)) )d~\tau.
		\end{aligned}
		\label{eq_a_2_1}
	\end{equation}
	The first step is to observe that cumulative cost is a suitable function to summarize the state of the learning at any time $t.$ The integral is indefinite and therefore represents a family of non-negative functions, parameterized by $\hat{\vect{\theta}}(\tau)$ and  $\vect{x}(t).$ The non-negative nature of the function is guaranteed by the construction of the cost and the choice of the cost. Furthermore, the function is zero when both  $\hat{\vect{\theta}}(\tau)$ and  $\vect{x}(t).$ are zero. The  function is continuously differentiable by construction. Lemma 1 describes boundedness  and existence of the convergence point for $J(\tau,  \hat{\vect{\theta}}(\tau)) )$ for all $\tau \in [t, \Gamma).$  We will also assume that there exists a bounded compact set~(search space) for $\hat{\vect{\theta}}(\tau)$~(the weight initialization procedure ensures this) and that input $x(t)$ is always numerically bounded~(this can be achieved through data normalization methods). With all these conditions being true, we can observe that  Eq.\ \ref{eq_a_2_1} is a reasonable candidate for this analysis \citep{athalye2015necessary} and also explains the behavior of the system completely.
	
	Note that we can write the first derivative of the cost function \textit{(this was derived as part of the HJB derivation)} under the assumption that the boundary condition for the optimal cost $V^*$ is $V$ such that
	\begin{equation}
		\begin{aligned}
			\frac{\partial V(t;\hat{\vect{\theta}}(t))}{\partial t}  &=  - \bigg[J(t;\hat{\vect{\theta}}(t)) )  +  \bigg( V_{\hat{\vect{\theta}}(t)} \bigg)^{T} \Delta \hat{\vect{\theta}}(t) & \\ & + \bigg( V_{x} \bigg)^{T} \Delta x(t)
			\bigg].&
		\end{aligned}
		\label{eq_a_2_2}
	\end{equation}
	Consider the update as $\Delta \hat{\vect{\theta}}(t) = \alpha V_{\hat{\vect{\theta}}(t)},$ with $\alpha >0$, and write by the fundamental theorem of calculus and chain rule
	\begin{equation}
		\begin{aligned}
			V_{\hat{\vect{\theta}}(t)} &=&  \frac{\partial}{\partial \hat{\vect{\theta}}(t) }\int_{\tau=t}^{\Gamma} J(\tau,  \hat{\vect{\theta}}(\tau)) )d~\tau \\ &=& - \frac{\partial}{\partial \hat{\vect{\theta}}(t) } \int_{\tau=\Gamma}^{t}  J(\tau, \vect{y}(\tau; \hat{\vect{\theta}}(\tau)) )d~\tau \\
			&=& - \frac{\partial}{\partial \hat{\vect{\theta}}(t) } [ \int_{\tau=\Gamma}^{c}  J(\tau, \vect{y}(\tau; \hat{\vect{\theta}}(\tau)) )d~\tau  \\ &+&  \int_{\tau=c}^{t}  J(\tau, \vect{y}(\tau; \hat{\vect{\theta}}(\tau)) )d~\tau ]\\
			&=& -  J(t, \vect{y}(t; \hat{\vect{\theta}}(t)) ) \frac{\partial J(t;\hat{\vect{\theta}}(t) }{\partial \hat{\vect{\theta}}(t) } \\
			&=& - J(t;\hat{\vect{\theta}}(t))  J_{\hat{\vect{\theta}}(t)}(t;\hat{\vect{\theta}}(t)) ) \\ &=& - J(t;\hat{\vect{\theta}}(t))  J_{\hat{\vect{\theta}}(t)}.
		\end{aligned}
		\label{eq_a_2_3}
	\end{equation}
	Similarly, simplify  $V_{x}$, and write by the fundamental theorem of calculus and chain rule
	\begin{equation}
		\begin{aligned}
			V_{x} &=&  \frac{\partial}{\partial x }\int_{\tau=t}^{\Gamma}  J(\tau,  \hat{\vect{\theta}}(\tau)) )d~\tau \\ &=& - \frac{\partial}{\partial x } \int_{\tau=\Gamma}^{t}  J(\tau, \vect{y}(\tau; \hat{\vect{\theta}}(\tau)) )d~\tau \\
			&=& \frac{\partial}{\partial x } [ \int_{\tau=\Gamma}^{c}  J(\tau, \vect{y}(\tau; \hat{\vect{\theta}}(\tau)) )d~\tau  \\ &+&  \int_{\tau=c}^{t}  J(\tau, \vect{y}(\tau; \hat{\vect{\theta}}(\tau)) )d~\tau ] \\
			&=& -  J(t, \vect{y}(t; \hat{\vect{\theta}}(t)) ) \frac{\partial J(t;\hat{\vect{\theta}}(t) }{\partial x } \\
			&=& - J(t;\hat{\vect{\theta}}(t))  J_{x}(t;\hat{\vect{\theta}}(t)) ) \\ &=& - J(t;\hat{\vect{\theta}}(t))  J_{x}.
		\end{aligned}
		\label{eq_a_2_4}
	\end{equation}
	
	Substituting Eqs.\ \eqref{eq_a_2_3} and \eqref{eq_a_2_4} into \eqref{eq_a_2_2}, we can write 
	\begin{equation}
		\begin{aligned}
			\frac{\partial V(t;\hat{\vect{\theta}}(t)) }{\partial t} & = - \bigg[  J(t;\hat{\vect{\theta}}(t)) )    & \\ &  -  \bigg(  J(t;\hat{\vect{\theta}}(t))  J_{\hat{\vect{\theta}}(t)} \bigg)^{T}(\alpha(t)  J(t;\hat{\vect{\theta}}(t))  J_{\hat{\vect{\theta}}(t)})    & \\ &  - \bigg(  J(t;\hat{\vect{\theta}}(t))  J_{x} \bigg)^{T} \Delta x(t)
			\bigg],
		\end{aligned}
		\label{eq_a_2_5}
	\end{equation}
	which when simplified provides
	\begin{equation}
		\begin{aligned}
			\frac{\partial V(t;\hat{\vect{\theta}}(t)) }{\partial t} & = - \bigg[ J(t;\hat{\vect{\theta}}(t)) )  - \alpha(t) J(t;\hat{\vect{\theta}}(t))^{2}  ||J_{\hat{\vect{\theta}}(t)}||^{2}   & \\ & -  J(t;\hat{\vect{\theta}}(t))  \bigg(J_{x}\bigg)^{T} \Delta x(t)
			\bigg].&
		\end{aligned}
		\label{eq_a_2_6}
	\end{equation}
	Pulling $J(t;\hat{\vect{\theta}}(t)) $ out of the bracket provides
	\begin{equation}
		\begin{aligned}
			\frac{\partial V(t;\hat{\vect{\theta}}(t))}{\partial t} & = -  J(t;\hat{\vect{\theta}}(t)) ) \bigg[1  - \alpha(t)  J(t;\hat{\vect{\theta}}(t)) ||J_{\hat{\vect{\theta}}(t)}||^{2} & \\ & - \bigg(J_{x}\bigg)^{T} \Delta x(t)
			\bigg].&
		\end{aligned}
		\label{eq_a_2_7}
	\end{equation}
	
	The first term $J(t;\hat{\vect{\theta}}(t)) )\geq 0$ and bounded by construction of the cost function. Equation \eqref{eq_a_2_7} is negative as long as the terms in the bracket are greater than or equal to zero, which is possible if and only if
	\begin{equation}
		\begin{aligned}
			\bigg[\alpha(t) J(t;\hat{\vect{\theta}}(t)) ||J_{\hat{\vect{\theta}}(t)}||^{2}+  \bigg(J_{x}\bigg)^{T} \Delta x(t)
			\bigg] < 1.
		\end{aligned}
		\label{eq_a_2_8}
	\end{equation}
	Hence, by Cauchy's inequality,  
	\begin{equation}
		\begin{aligned}
			\bigg[\alpha J(t;\hat{\vect{\theta}}(t)) ||J_{\hat{\vect{\theta}}(t)}||^{2}+   \|J_{x}\| \|\Delta x(t)\| \bigg] < 1, \\
			J(t;\hat{\vect{\theta}}(t)) < \frac{1 - \|J_{x}\| \|\Delta x(t)\| }{\alpha(t) ||J_{\hat{\vect{\theta}}(t)}||^{2} } .
		\end{aligned}
		\label{eq_a_2_10}
	\end{equation}
	Choose $\alpha(t)  = \frac{1 - \|J_{x}\| \|\Delta x(t)\| }{\beta ||J_{\hat{\vect{\theta}}(t)}||^2}$ with $\beta>0$, and get the bound on $ \| J(t;\hat{\vect{\theta}}(t))\|$ as
	\begin{equation}
		\begin{aligned}
			J(t;\hat{\vect{\theta}}(t)) < \beta .
		\end{aligned}
		\label{eq_a_2_11}
	\end{equation}
	As a consequence, $V(t;\hat{\vect{\theta}}(t))$ is ultimately bounded \citep{lewis2012optimal} with  $\| J(t;\hat{\vect{\theta}}(t))\| < \beta. $
\end{proof}
\subsection{Derivation of the Update through Finite Approximation}
From Theorem 1, the update for the network is chosen as the derivative of the cumulative cost, that is, $\frac{\partial V(t, \vect{y} (t; \hat{\vect{\theta}}(t)))}{\partial \hat{\vect{\theta}}(t) }$ providing
\begin{equation}
	\begin{aligned}
		& - \frac{\partial V^{*}(t; \hat{\vect{\theta}}(t))}{\partial t } = min_{\hat{\vect{\theta}}(t) \in \Omega}  \bigg[ H(t;  \hat{\vect{\theta}}(t)))  \bigg],&
	\end{aligned}
	\label{eq_3_1}
\end{equation}
where $H(t;  \hat{\vect{\theta}}(t))) $ is the CT Hamiltonian, which we will discretize and approximate. Under the assumption that the boundary condition for the optimal cost is the cumulative cost itself, we may write
\begin{equation}
	\begin{aligned}
		&H(t;  \hat{\vect{\theta}}(t))) = J(t;\hat{\vect{\theta}}(t))& \\ &+\big(  V_{\hat{\vect{\theta}}(t)}\big)^{T} \Delta \hat{\vect{\theta}}  + \big( V_{\vect{x}(t)} \big)^{T} \Delta \vect{x}(t).&
	\end{aligned}
	\label{eq_3_2}
\end{equation}
Upon Euler's disretization, we achieve 
\begin{equation}
	\begin{aligned}
		& \frac{1}{\Delta t} H(k;  \hat{\vect{\theta}}(k))) = \frac{1}{\Delta t} J(k;\hat{\vect{\theta}}(k))& \\ &+\big( V(k; \hat{\vect{\theta}}(k+1))- V(k; \hat{\vect{\theta}}(k)) & \\ & +\big( V(k+1; \hat{\vect{\theta}}(k))- V(k; \hat{\vect{\theta}}(k)) &
	\end{aligned}
	\label{eq_3_3}
\end{equation}
Simplification provides 
\begin{equation}
	\begin{aligned}
		& H(k;  \hat{\vect{\theta}}(k))) = J(k;\hat{\vect{\theta}}(k))& \\ &+ \Delta t \big( V(k; \hat{\vect{\theta}}(k+1))- V(k; \hat{\vect{\theta}}(k)) & \\ & + \Delta t \big( V(k+1; \hat{\vect{\theta}}(k))- V^*(k; \hat{\vect{\theta}}(k)) &
	\end{aligned}
	\label{eq_3_4}
\end{equation}
Taking the derivative and setting it to zero, we get 
\begin{equation}
	\begin{aligned}
		& 0 = \frac{\partial}{\partial \hat{\vect{\theta}}(k) }~J(k;\hat{\vect{\theta}}(k))& \\ &+ \frac{\partial}{\partial \hat{\vect{\theta}}(k) }~\Delta t \big( V(k; \hat{\vect{\theta}}(k+1))- V(k; \hat{\vect{\theta}}(k)) & \\ & + \frac{\partial}{\partial \hat{\vect{\theta}}(k) }~ \Delta t \big( V(k+1; \hat{\vect{\theta}}(k))- V(k; \hat{\vect{\theta}}(k)).&
	\end{aligned}
	\label{eq_3_5}
\end{equation}
Rearranging, we  get 
\begin{equation}
	\begin{aligned}
		& \frac{\partial V(k; \hat{\vect{\theta}}(k))  }{\partial \hat{\vect{\theta}}(k) } = \bigg[ \frac{\partial}{\partial \hat{\vect{\theta}}(k) }~\frac{J(k;\hat{\vect{\theta}}(k))}{\Delta t}  & \\ &  + \frac{\partial}{\partial \hat{\vect{\theta}}(k) } \big( V(k; \hat{\vect{\theta}}(k+1)))  - V(k; \hat{\vect{\theta}}(k)) \big)& \\ &+ \frac{\partial}{\partial \hat{\vect{\theta}}(k)}V(k+1; \hat{\vect{\theta}}(k))\bigg].&
	\end{aligned}
	\label{eq_3_6}
\end{equation}
Since we have no information about the data from the future, we will think of the last term $ \frac{\partial}{\partial \hat{\vect{\theta}}(k) }\big( V(k+1; \hat{\vect{\theta}}(k))$ as $0$.
Under the assumption that $  \frac{\partial}{\partial \hat{\vect{\theta}}(t) } \int_{\tau=t}^{c} J(\tau,  \hat{\vect{\theta}}(\tau)) )d~\tau~\propto~J(t,\hat{\vect{\theta}}(t)) ).$ From the fundamental theorem of calculus we write
\begin{equation}
	\begin{aligned}
		V_{\hat{\vect{\theta}}(t)} &= \frac{\partial}{\partial \hat{\vect{\theta}}(t) }\int_{\tau=t}^{\Gamma} J(\tau,  \hat{\vect{\theta}}(\tau)) )d~\tau& \\ &= - \frac{\partial}{\partial \hat{\vect{\theta}}(t) } \int_{\tau=\Gamma}^{t}  J(\tau, \vect{y}(\tau; \hat{\vect{\theta}}(\tau)) )d~\tau& \\
		&\propto -  J(t; \hat{\vect{\theta}}(t)) \frac{\partial J(t;\hat{\vect{\theta}}(t) }{\partial \hat{\vect{\theta}}(t) } &\\
		&\propto -  J(t;\hat{\vect{\theta}}(t)) ). &
	\end{aligned}
	\label{eq_3_7}
\end{equation}
We now achieve our update as
\begin{equation}
	\begin{aligned}
		& \frac{\partial V(k; \hat{\vect{\theta}}(k))  }{\partial \hat{\vect{\theta}}(k) } \propto  \bigg[ \frac{\partial}{\partial \hat{\vect{\theta}}(k) }~\frac{J(k;\hat{\vect{\theta}}(k))}{\Delta t}  & \\ &  - \frac{\partial}{\partial \hat{\vect{\theta}}(k) } \big( J(k; \hat{\vect{\theta}}(k+1)) - J(k; \hat{\vect{\theta}}(k))) \big) \bigg] .&
	\end{aligned}
	\label{eq_3_8}
\end{equation}
To obtain the first term, we  replace $1/\Delta t$ with a small value $\eta$ and write $J(k; \hat{\vect{\theta}}(k)). = J_{N}(k; \hat{\vect{\theta}}(k)) + J_{P}(k; \hat{\vect{\theta}}(k)),$ where $J_{P}(k; \hat{\vect{\theta}}(k))$ refers to the cost on all the previous tasks and  $J_{N}(k; \hat{\vect{\theta}}(k))$ refers to the cost on the new task. For the second term, we  replace the difference  $\big( J(k; \hat{\vect{\theta}}(k+1)))  - J(k; \hat{\vect{\theta}}(k)) \big)$ with $\big( J(k; \hat{\vect{\theta}}(k+\zeta)))  - J(k; \hat{\vect{\theta}}(k)) \big)$ with $\zeta$ being the finite number of steps for the approximation.
We get the gradient 
\begin{equation}
	\begin{aligned}
		& \frac{\partial V(k; \hat{\vect{\theta}}(k))  }{\partial \hat{\vect{\theta}}(k) } \propto  \bigg[\frac{\partial}{\partial \hat{\vect{\theta}}(k) }~\eta[J_{N}(k; \hat{\vect{\theta}}(k)) +  J_{P}(k; \hat{\vect{\theta}}(k))]  & \\ &  + \frac{\partial}{\partial \hat{\vect{\theta}}(k) } \big(J(k; \hat{\vect{\theta}}(k)) - J(k; \hat{\vect{\theta}}(k+\zeta)))\big),\bigg],&
	\end{aligned}
	\label{eq_PDE}
\end{equation}
where $\zeta$ is a predefined number of iterations for the parameters.  To simplify notation, we  write $J_{N}(k; \hat{\vect{\theta}}(k))$ as $J_{N}(k),$ $J_{N}(k; \hat{\vect{\theta}}(k))$ as $J_{P}(k),$
$J(k; \hat{\vect{\theta}}(k))$  as $J_{PN}(k)$, and 
$J(k; \hat{\vect{\theta}}(k+\zeta))$  as $J_{PN}(k; \hat{\vect{\theta}}(k+\zeta))$ and 
\begin{equation}
	\begin{aligned}
		& \frac{\partial V(k; \hat{\vect{\theta}}(k))  }{\partial \hat{\vect{\theta}}(k) } \propto  \bigg[\frac{\partial}{\partial \hat{\vect{\theta}}(k) }~\eta[J_{N}(k) +  J_{P}(k)]  & \\ &  + \frac{\partial}{\partial \hat{\vect{\theta}}(k) } \big(  J_{PN}(k) - J_{PN}(k; \hat{\vect{\theta}}(k+\zeta))) \big),\bigg]. .&
	\end{aligned}
\end{equation}
Our update rule  with $\eta = 1$ then is
\begin{equation}
	\begin{aligned}
		\hat{\vect{\theta}}(k+1)   &= \hat{\vect{\theta}}(k) -  \alpha(t) \times \bigg[\frac{\partial}{\partial \hat{\vect{\theta}}(k) }~[J_{N}(k) +  J_{P}(k)]  & \\ &  + \frac{\partial}{\partial \hat{\vect{\theta}}(k) } \big( J_{PN}(k) - J_{PN}(k; \hat{\vect{\theta}}(k+\zeta))) \big),\bigg].&
	\end{aligned}
	\label{eq4}
\end{equation}
\section{Appendix -  Results and Implementation Details}
Our implementations are done in Python with the Pytorch library (version 1.4). First, we will discuss the datasets

\subsection{Data set}
\noindent \textit{Incremental Sine Waves} (regression problem): An incremental sine wave problem is defined by fifty (randomly generated) sine functions where each sine wave is considered a task and is incrementally shown to the model. Each sine function is generated by randomly selecting an amplitude in the range $[0.1, 5]$ and phase in $[0, \pi]$. For  training, we generate 40 minibatches from the first sine function in the sequence (each minibatch has eight elements) and then 40 from the second and so on.  We use a single regression head to predict these tasks.  For the time, $t \in \{0, 0.001, \cdots, 0.01\}.$ We generate a sine wave data set consisting of 50 tasks. Each task is shown sequentially to the model and is described by its amplitude, phase, frequency, and  time $t \in \{0, 0.001, \cdots, 0.01\}$. Each task is generated by making incremental changes to amplitude, phase, and frequency while following the protocol in \citep{finn2017model}.

\noindent \textit{Split-Omniglot data set):}  We choose the first fifty classes to constitute our problem. Each character has 20 handwritten images. The data set is divided into two parts. These classes are shown incrementally to all the approaches. We choose  12 images to be part of the training data for each task, 3 images for validation, and   5 images for the testing.

\noindent \textit{MNIST \& CIFAR10 data set):}  These data-sets are comprised of ten classes. We incremently show each of these classes to our model. Therefore, each task in our problem is comprised of exactly one class. We choose 60 \% of the data from each task to constitute our training data and the rest is split into validation and test.

\subsection{Model Setup}
\textit{SINE:}  DPMCL uses two neural networks: one for the representation and one for the prediction. Both have a three-layer feed-forward network (one input, one hidden, and one output layer)  with $100$ hidden layer neurons and $relu$ activation function.The input vector is $3\times 1$m and the output of the prediction network is a $100 \times 1$ vector with a linear activation function at the output layer.  For the implementations of Naive, ER, and OML, a single six-layer network (100 hidden units, relu activation) is utilized. For CML, we use two networks: representation learning network and prediction learning networks. For ANML, the representation learning network becomes the neuromodulatory network \citep{beaulieu2020learning}. For CML and ANML implementations, we use 100 hidden units (same as DPMCL). 

\subsubsection{Sine Model Definitions}
\begin{lstlisting}
    # Model Feature extractions.
    self.model_F = torch.nn.Sequential(
        torch.nn.Linear(self.config['D_in'], self.config['H']),
        torch.nn.ReLU(),
        torch.nn.Linear(self.config['H'], self.config['H']),
        torch.nn.ReLU(),
        torch.nn.Linear(self.config['H'], self.config['D_in'])
    )

    # The g model and the buffer model are the same
    # Model g
    self.model_P = torch.nn.Sequential(
    torch.nn.Linear(self.config['D_in'], self.config['H']),
    torch.nn.ReLU(),
    torch.nn.Linear(self.config['H'], self.config['H']),
    torch.nn.ReLU(),
    torch.nn.Linear(self.config['H'], self.config['D_out']) 
    )


    # Model buffer
    self.model_buffer = torch.nn.Sequential(
    torch.nn.Linear(self.config['D_in'], self.config['H']),
    torch.nn.ReLU(),
    torch.nn.Linear(self.config['H'], self.config['H']),
    torch.nn.ReLU(),
    torch.nn.Linear(self.config['H'], self.config['D_out']) 
    )
\end{lstlisting}

\textit{MNIST, OMNI, CIFAR10:} For all these data sets, we use a combination of  convolutional neural network (two convolutional layers, max pooling, relu-activation function) and a feed-forward network (2 feed-forward layer, relu activation function, softmax output). For CIFAR10, we use three channels, of which  only one channel is used with OMNI and MNIST. For the implementations of Naive, ER, and OML, a single four-layer network (2 convolutional layers, 2 feed-forward layers) is utilized. For CML \citep{javed2019meta}, we use two networks: a representation learning network (convolutional neural network) and a prediction learning network (feed-forward network), respectively. For ANML, the representation learning network becomes the neuromodulatory network \citep{beaulieu2020learning}.  

\subsubsection{ OMNI}
\begin{lstlisting}
    # Feature Extractor
    self.model_F = torch.nn.Sequential( 
    torch.nn.Conv2d(1, 32, kernel_size=5, stride=1, padding=2),
    torch.nn.MaxPool2d(kernel_size=2, stride=2),
    torch.nn.ReLU(),
    torch.nn.Conv2d(32, 64, kernel_size=5, stride=1, padding=2),
    torch.nn.MaxPool2d(kernel_size=2, stride=2),
    torch.nn.ReLU(),
    )

    # Feedforward Layer
    self.model_P = torch.nn.Sequential( 
    torch.nn.Linear(7 * 7 * 64, self.config['H']),
    torch.nn.ReLU(),
    torch.nn.Linear(self.config['H'],  self.config['D_out'])
    )
    
    # Buffer Layer
    self.model_buffer = torch.nn.Sequential( 
    torch.nn.Linear(7 * 7 * 64, self.config['H']),
    torch.nn.ReLU(),
    torch.nn.Linear(self.config['H'],  self.config['D_out'])
    )        
\end{lstlisting}

 \subsubsection{Cifar10 and MNIST}
 \begin{lstlisting}
 # Feature Extractor
      self.model_F = torch.nn.Sequential( 
    torch.nn.Conv2d(3, 6, 5),
    torch.nn.MaxPool2d(kernel_size=2, stride=2),
    torch.nn.ReLU(),
    torch.nn.Conv2d(6, 16, 5),
    torch.nn.MaxPool2d(kernel_size=2, stride=2),
    torch.nn.ReLU(),
    torch.nn.Dropout()
    )

# Feed Forward Layer
  self.model_P = torch.nn.Sequential( 
    torch.nn.Linear(256, self.config['H']),
    torch.nn.ReLU(),
    torch.nn.Linear(self.config['H'], self.config['H']),
    torch.nn.ReLU(),
    torch.nn.Linear(self.config['H'], self.config['D_out'])
    )
#Buffer Layer
  self.model_buffer = torch.nn.Sequential( 
    torch.nn.Linear(256, self.config['H']),
    torch.nn.ReLU(),
    torch.nn.Linear(self.config['H'], self.config['H']),
    torch.nn.ReLU(),
    torch.nn.Linear(self.config['H'], self.config['D_out'])
    )
 \end{lstlisting}

\subsection{Hyperparameters}
Since, DPMCL update rule involves division by the norm of the gradient, we use the Adagrad optimizer throughout.
\begin{table}[H]
	\centering
	\scriptsize{
	\begin{tabular}{l|l|l|l|l}
		\hline
		Parameters              & Sine                   & Omniglot              & MNIST                   & CIFAR10\\
		\hline
		Learning rate            & $1e-03$               &  $1e-04$              & $1e-04$                 &   $1e-04$           \\
		total runs               & $50$                  &  $50$                 & $50$                    &   $50$           \\
		num tasks                & $50$                  &  $50$                 & $10$                    &   $10$            \\  
		Num of Hidden Layers.    & $100$                 &  $100$                & $512$                   &   $512$             \\
		Input size               & $3$                   &  $28\times 28$        & $28 \times 28$          &   $28 \times 28$  \\    
		Output Size              & $100$                 &  $50$                 & $10$                    &   $10$\\
		$\kappa$                 & $300$                 &  $200$                & $300$                   &   $600$\\
		$\zeta$                  & $2$                   &  $2$                  & $5$                     &   $5$\\
		$N_{meta}$               & $150$                 &  $100$                & $150$                   &  $300$\\
		$N_{grad}$               & $150$                 &  $100$                & $150$                   &  $300$\\
		$N$                      & $300$                 &  $200$                & $300$                   &  $600$\\
		length of $D_{P}$        & $1000$                &  $20000$              & $20000$                 & $100000$\\
		$\beta$                  & $1000$                &  $10000$              & $10000$                 &  $10000$\\
        batch size		         & $64$                  &  $8$                  & $32$                    &  $512$\\
		activation function      & relu, output-linear   & relu,output-softmax  & relu, output-softmax    & relu, output-softmax\\
		optimizer                & Adagrad               & Adagrad              & Adagrad                 & Adagrad \\
		Loss function            & MSE                   & Cross Entropy        & Cross Entropy           & Cross Entropy\\
		\hline
	\end{tabular}}
\end{table}
Next, we will discuss the different methods that have been used in the study. We start by describing the algorithm for DPMCL.
\subsection{DPMCL}
We define a new task sample, $ \cD_N(k) = \{ \mathcal{X}_{k}, \mathcal{Y}_{k} \},$ and a task memory (samples from all the previous tasks) $\cD_{P}(k) \subset \cup_{\tau = 0}^{k-1}\cT_{\tau}$. We can approximate the required terms in our update rule Eq.~\ref{eq_update} using samples (batches) from $\cD_{P}(k)$ and $\cD_N(k).$  The overall algorithm consists of two steps: generalization  and  catastrophic forgetting~(see Algorithm 1 in Appendix C). DPMCL comprises representation and prediction neural networks parameterized by $\hat{\vect{\theta}}_{1}$ and $\hat{\vect{\theta}}_2$, respectively. For each batch $b_{N} \in \cD_N(k)$, DPMCL alternatively performs generalization and catastrophic forgetting cost updates $\kappa$ times. The generalization cost update consists of computing the cost $J_N$ and using that to update $\hat{\vect{\theta}}_{1}$ and $\hat{\vect{\theta}}_2$; the catastrophic forgetting cost update comprises the following steps. First we create a batch that combines the new task data with samples from the previous tasks $b_{PN} = b_{P} \cup b_{N}(k)$, where $b_{P} \in \cD_{P}(k)$. Second, to approximate the term $\big( J_{PN}(k; \hat{\vect{\theta}}(k+\zeta))),$,  we  copy $\hat{\vect{\theta}}_{2}$~(prediction network) into a temporary network parameterized by $\hat{\vect{\theta}}_{B}$. We  then perform $\zeta$ updates on $\hat{\vect{\theta}}_{B}$ while keeping $\hat{\vect{\theta}}_{1}$ fixed. Third, using $\hat{\vect{\theta}}_{B}(k+\zeta),$ we compute $J_{PN}(k;\hat{\vect{\theta}}_{B}(k+\zeta))$ and update $\hat{\vect{\theta}}_{1}, \hat{\vect{\theta}}_2$ with $J_{P}(k)+( J_{PN}(k) - J_{PN}(k;\hat{\vect{\theta}}_{B}(k+\zeta)) )$~(lines 16-17 in Alg.~1).

\begin{algorithm}[H]
	\SetAlgoLined
	Initialize $\hat{\vect{\theta}}_{1},\hat{\vect{\theta}}_{2}, D_{P}, D_{N}$ \\
	\While{$k=1,2,3,... k \times \Gamma$}{
		i = 0
		\While{$i < \kappa$}{
			\textbf{Step 1: Generalization}
			Get $\vect{b}_{N} \in D_{N}(k)$
			Update~$\hat{\vect{\theta}}_{1}(k),\hat{\vect{\theta}}_{2}(k)$ with $J_{N}(k)$   \\
			\textbf{Step 2: Catastrophic Forgetting} 
			Get $J_{P}(k)$ with $\vect{b}_{P} \in D_{P}(k)$
			Get $\vect{b}_{PN} = \vect{b}_{P} \cup \vect{b}_{N}$ and copy $\hat{\vect{\theta}}_{2}$ into $\hat{\vect{\theta}}_{B}$ \\
			j = 0\\
			\While{$j+1 <= \zeta$}{
				Update~$\hat{\vect{\theta}}_{B}(k)$ with $J_{PN}(k;\hat{\vect{\theta}}_{B}(k)).$\\
				j = j+1 }
			Update~$\hat{\vect{\theta}}_{2}(k),\hat{\vect{\theta}}_{1}(k)$ with $J_{P}(k)+(J_{PN}(k) - J_{PN}(k;\hat{\vect{\theta}}_{B}(k+\zeta))).$	\\
			i = i+1
		}	
		Update $D_{P}(k)$ with  $D_{N}(k).$}
\end{algorithm}

\subsection{Comparative methods}
The five methods are Naive, ER, OML, CML and ANML.
\textbf{Naive}
For the naive implementation, we use the training data for each task  to train our approach. The core idea is to greedily learn  any new task. We run gradient updates for each task data for a predetermined number of epochs.
\begin{algorithm}[H]
	\SetAlgoLined
	Initialize $\vect{\theta}(k).$ \\
	\While{$j< num~tasks$}{
		Initialize task data $D_N$ \\
		k =0\\
		\While{$k< N$}{
			$b_{N} \in D_{N}$\\
			Update $\vect{\theta}$\\
		k = k+1\\
		}
	}
	\caption{Naive algorithm.}
	\label{alg_Naive}
\end{algorithm}

\textbf{Experience-Replay (ER~\citep{lin1992self}):} This approach aims at maintaining the performance of all the tasks till now. We therefore define a task memory array. We store samples from each new task into the experience replay array. At the start of every new task, we use the samples from the task memory array for training the network multiple epochs through the data. This method focuses on minimizing catastrophic forgetting.
\begin{algorithm}[H]
	\SetAlgoLined
	Initialize $\vect{\theta}(k)$ and $D_{PN}$\\
	\While{$j< num~tasks$}{
		Initialize task data $D_N$ and append to $D_{PN}$\\
		k =0\\
		\While{$k< N$}{
			$b_{PN} \in D_{PN}$\\
			Update $\vect{\theta}$\\
		k = k+1\\
		}
	}
	\caption{Experience Replay Algorithm.}
	\label{alg_ER}
\end{algorithm}

\textbf{Online Meta Learning (OML,~\citep{finn2019online}):}  We follow the meta training testing procedures described in \citep{finn2019online} for this implementation. The process is composed of two loops. In the inner loop, the training is performed on the new task;  in the outer loop, the training is performed on the buffer (task memory). We first save samples from each task into the buffer data. Both the inner loop and the outer loop updates are performed by using the gradients of the cost function.
\begin{algorithm}[H]
	\SetAlgoLined
	Initialize $\vect{\theta}(t).$ \\
	Initialize $D_{PN}$ \\
	\While{$j< num~tasks$}{
		Initialize $D_{N}$ and append to $D_{PN}$
		\For{$N_{meta}$ samples in $D_{PN}$}{Update $\vect{\theta}(k)$ to obtain $\tilde{\vect{\theta}}(k)$}
		\For{$N_{grad}$ samples in $D_N$}{Update $\vect{\theta}(k)$ using cost calculated with $\tilde{\vect{\theta}}.$}
	}
	\caption{Our OML implementation.}
\end{algorithm}

\textbf{Online Meta Continual Learning (CML,~\citep{javed2019meta}):} The learning process of this method is the same as that of the one in \citep{finn2019online} with the key difference being the use of the representation network. The algorithm is provided in \citep{javed2019meta}. This approach is composed of a prelearned representation. We do not train a representation but try to learn it while the tasks are being observed sequentially. This protocol is followed to highlight the idea that although good representations are necessary,  no data is available for training a representation. 
\begin{algorithm}[H]
    \SetAlgoLined
	Initialize $\vect{\theta}_{1}(t), \vect{\theta}_{2}(t).$ \\
	Initialize $D_{PN}$ \\
	\While{$j< num~tasks$}{
		Initialize $D_{N}$ and append to $D_{PN}$
		\For{$N_{meta}$ samples in $D_{PN}$}{
		Update $\vect{\theta}_{2}(k)$ to obtain $\tilde{\vect{\theta}}_{2}(k)$\\ 
		}
		\For{$N_{grad}$ samples in $D_N$}{Update $\vect{\theta}(k)$ using cost calculated with $\tilde{\vect{\theta}}_{2}$ and $\vect{\theta}_{1}$ }
	}
	\caption{Our CML Implementation.}
	\label{alg_CML}
\end{algorithm}
\textbf{Neuromodulated Meta Learning (ANML~\citep{beaulieu2020learning}):}  Similar to the CML case, the learning process is the same as that of OML with the key difference being the neuromodulatory network which is in addition to the representation learning network. The algorithm is provided in \citep{beaulieu2020learning}. Similar to the earlier scenario, a the neuromodulatory network is learned while the tasks are being observed.
\begin{algorithm}[H]
    \SetAlgoLined
	Initialize Network 1 with parameters $\vect{\theta}_{1}(t)$\\
	Initialize Network 2 with $\vect{\theta}_{2}(t)$ \\
	Initialize $D_{PN}$ \\
	\While{$j< num~tasks$}{
		Initialize $D_{N}$ and append to $D_{PN}$\\
		\For{$N_{meta}$ samples in $D_{PN}$}{
		Get output by multiplying Network 1 and Network 2 outputs(similar to gating process in \citet{beaulieu2020learning}).\\
		Update $\vect{\theta}_{2}(k)$\\ 
		}
		\For{$N_{grad}$ samples in $D_N$}{
		Get output by multiplying Network 1 and Network 2 outputs(similar to gating process in \citet{beaulieu2020learning}).\\
		Update $\vect{\theta}_{1}(k)$ using cost calculated with $\vect{\theta}_{2}$ and $\vect{\theta}_{1}$ }
	}
	\caption{Our CML Implementation.}
	\label{alg_CML}
\end{algorithm}

\subsection{Additional Results}
\begin{table}[H]
    \footnotesize{
	\centering
	\begin{tabular}{ll|llllll}
		\hline
		                &           & Naive              &\textbf{DPMCL}   &OML            & CML                 & ANML                  & ER \\
		\hline
		\multirow{2}{*}{SINE}       &CME& $10^{-4}$(0)    & $10^{-5}$(0)        & $10^{-5}$(0)    & $0.0005$(0)       & $10^{-4}$(0)      & $10^{-5}$(0)     \\
		                            &NTE& $10^{-7}$(0)    & $10^{-7}$(0)         & $10^{-05}$(0)  & $10^{-7}$(0)       & $10^{-7}$(7)      &  $10^{-5}$(0)  \\
		                            \hline
		\multirow{2}{*}{OMNI}       &CME& 0.979(0)        & 0.171(0.007)       & 0.224(0.010)    & 0.706(0.245)       & 0.977(0.001)      & 0.194(0.008)     \\
		                            &NTE& 0.001(0)        & 0.283(0.091)       & 0.512(0.098)    & 0.077(0.053)       & 0.945(0)          & 0.291(0.071)     \\
		                            \hline
		\multirow{2}{*}{MNIST}      &CME& 0.912(0)        & 0.020(0.001)       & 0.023(0.001)    & 0.659(0.004)       & 0.873(0.002)      & 0.030(0.001)     \\
		                            &NTE& 0.001(0)        & 0.003(0)           & 0.011(0)        & 0.001(0)       & 0.884(0)          & 0.024(0.001)     \\
		                            \hline
		\multirow{2}{*}{CIFAR10}    &CME& 0.949(2)        & 0.496(0.003)       & 0.676(0.006)    & 0.651(0.004)       & 0.918(0.016)      & 0.464(0.002)     \\
		                            &NTE& 0.01(0)         & 0.231(0.008)       & 0.108(0.005)    & 0.055(0.003)       & 0.689(0.140)      & 0.458(0.008)     \\
		\hline
	\end{tabular}
	\caption{cumulative error~(CME) and new task error~(NTE)value for different data sets. The mean and standard error of the mean are reported. These values are calculated by averaging across 50 repetitions\label{table:table_CME}.}}
\end{table}

\bibliography{cdc.bib}

\end{document}